\documentclass{article} % For LaTeX2e
\usepackage{iclr2025_conference,times}

\usepackage{amsmath,amsfonts,bm}

\def\eqref#1{equation~\ref{#1}}
\def\1{\bm{1}}

\DeclareMathAlphabet{\mathsfit}{\encodingdefault}{\sfdefault}{m}{sl}
\SetMathAlphabet{\mathsfit}{bold}{\encodingdefault}{\sfdefault}{bx}{n}

\usepackage{hyperref}
\usepackage{url}

\title{X-ALMA: Plug \& Play Modules and Adaptive Rejection for Quality Translation at Scale}

\stepcounter{footnote}\stepcounter{footnote}
\author{%
Haoran Xu$^{\circ}$ Kenton Murray$^{\triangleleft}$ Philipp Koehn$^{\triangleleft}$ Hieu Hoang$^{\circ}$ Akiko I. Eriguchi$^{\circ}$ Huda Khayrallah$^{\diamond,}$\thanks{Work done at Microsoft.}  \\
$^{\circ}$Microsoft, $^{\triangleleft}$Johns Hopkins University, $^{\diamond}$Amazon\\
\texttt{haoranxu@microsoft.com},
\texttt{$\{$kenton, phi$\}$@jhu.edu},\\\texttt{$\{$hihoan,akikoe$\}$@microsoft.com}, \texttt{hudakh@amazon.com}\\[0.2em]
}

\usepackage{amsmath}
\usepackage{amssymb}
\usepackage{mathtools}
\usepackage{amsthm}

\usepackage{longtable}
\usepackage{wrapfig}
\usepackage{url}
\usepackage{color}
\usepackage{makecell}
\usepackage{bbm}
\usepackage{amsmath}
\usepackage{arydshln}
\usepackage{multirow}
\usepackage{booktabs}
\usepackage{amssymb}
\usepackage{xcolor,pifont}
\usepackage{CJKutf8}

\definecolor{mygreen}{HTML}{0F911C}
\definecolor{myred}{HTML}{E92121}
\newcommand{\green}[1]{\textcolor{mygreen}{#1}}
\newcommand{\red}[1]{\textcolor{myred}{#1}}

\usepackage{graphicx,calc}
\newlength\myheight
\newlength\mydepth
\settototalheight\myheight{Xygp}
\newcommand*\inlinegraphics[1]{%
  \settototalheight\myheight{Xygp}%
  \settodepth\mydepth{Xygp}%
  \raisebox{-\mydepth}{\includegraphics[height=\myheight]{#1}}%
}

\newcommand\extrafootertext[1]{%
    \bgroup
    \renewcommand\thefootnote{\fnsymbol{footnote}}%
    \renewcommand\thempfootnote{\fnsymbol{mpfootnote}}%
    \footnotetext[0]{#1}%
    \egroup
}

\iclrfinalcopy % Uncomment for camera-ready version, but NOT for submission.
\begin{document}

\maketitle
% \extrafootertext{Work done during an internship at Microsoft.}
\begin{abstract}
% \textcolor{purple}{I am trying to make the authors all work w/o taking up a full half page. if there are strong objections, feel free to change or let me know - hk}

Large language models (LLMs) have achieved remarkable success across various NLP tasks with a focus on English due to English-centric pre-training and limited multilingual data. In this work, we focus on the problem of translation, and 
while some multilingual LLMs claim to support for hundreds of languages, models often fail to provide high-quality responses for mid- and low-resource languages, leading to imbalanced performance heavily skewed in favor of high-resource languages. We introduce \textbf{X-ALMA}, a model designed to ensure top-tier performance across 50 diverse languages, regardless of their resource levels. X-ALMA surpasses state-of-the-art open-source multilingual LLMs, such as Aya-101 \citep{aya101} and Aya-23 \citep{aya23}, in every single translation direction on the FLORES-200 and WMT'23 test datasets according to COMET-22. This is achieved by plug-and-play language-specific module architecture to prevent language conflicts during training and a carefully designed training regimen with novel optimization methods to maximize the translation performance. After the final stage of training regimen, our proposed \underline{\textbf{A}}daptive-\underline{\textbf{R}}ejection \underline{\textbf{P}}reference \underline{\textbf{O}}ptimization (\textbf{ARPO}) surpasses existing preference optimization methods in translation tasks.\footnote{Code is released at \url{https://github.com/fe1ixxu/ALMA}. Models and Dataset are released at \href{https://huggingface.co/collections/haoranxu/x-alma-66fde464ef90be465920abaa}{\texttt{https://huggingface/X-ALMA}}.}
\end{abstract}

\section{Introduction}
Large language models (LLMs) such as the GPT series \citep{gpt3_few_shot,openai2023gpt4}, Mistral \citep{mistral}, LLaMA series \citep{llama1,llama2,llama3}, Gemma series \citep{gemma1,gemma2}, 
\textit{inter alia}
have demonstrated impressive performance across various NLP tasks. However, the efficacy of LLMs has primarily been evaluated on English tasks, with their multilingual capabilities receiving less attention due to the models being predominantly pre-trained on English and the scarcity of multilingual data. Recently, there has been a shift towards multilingual studies in LLMs. For instance, LLaMA 3 and 3.1 \citep{llama3} expand the vocabulary from 32K to 128K and pre-train on multilingual texts; \citet{aya101} have introduced Aya-101, a multilingual generative model supporting 101 languages; and BigTranslate \citep{bigtranslate} and LLaMAX \citep{llamax} scale LLM-based multilingual translation models to over 100 languages.

Despite the increased language support in LLMs, their performance across most languages falls short of practical application expectations, especially for mid- and low-resource languages (\textit{weakness 1}). Furthermore, the performance of high-resource languages tends to be inferior compared to LLMs trained with fewer languages, a phenomenon known as the `curse of multilinguality' \citep{xlmr} (\textit{weakness 2}). The weaknesses are prevalent in most current state-of-the-art (SoTA) massively multilingual models: overall quality decreases as the number of supported languages increases. Although methods such as building models by focusing on a smaller number of high-resource languages like German and Chinese can achieve satisfactory performance for these languages and mitigate these weaknesses \citep{aya23,alma,alma-r,tower}, they neglect the needs of mid- and low-resource languages. In this work, we address these weaknesses and build a multilingual model that achieves consistently high performance across 50 diverse languages, regardless of resource level, with a focus on multilingual machine translation.

\begin{figure}[ht]
    \centering
    \resizebox{0.8\linewidth}{!}{
    \includegraphics[width=7.5cm]{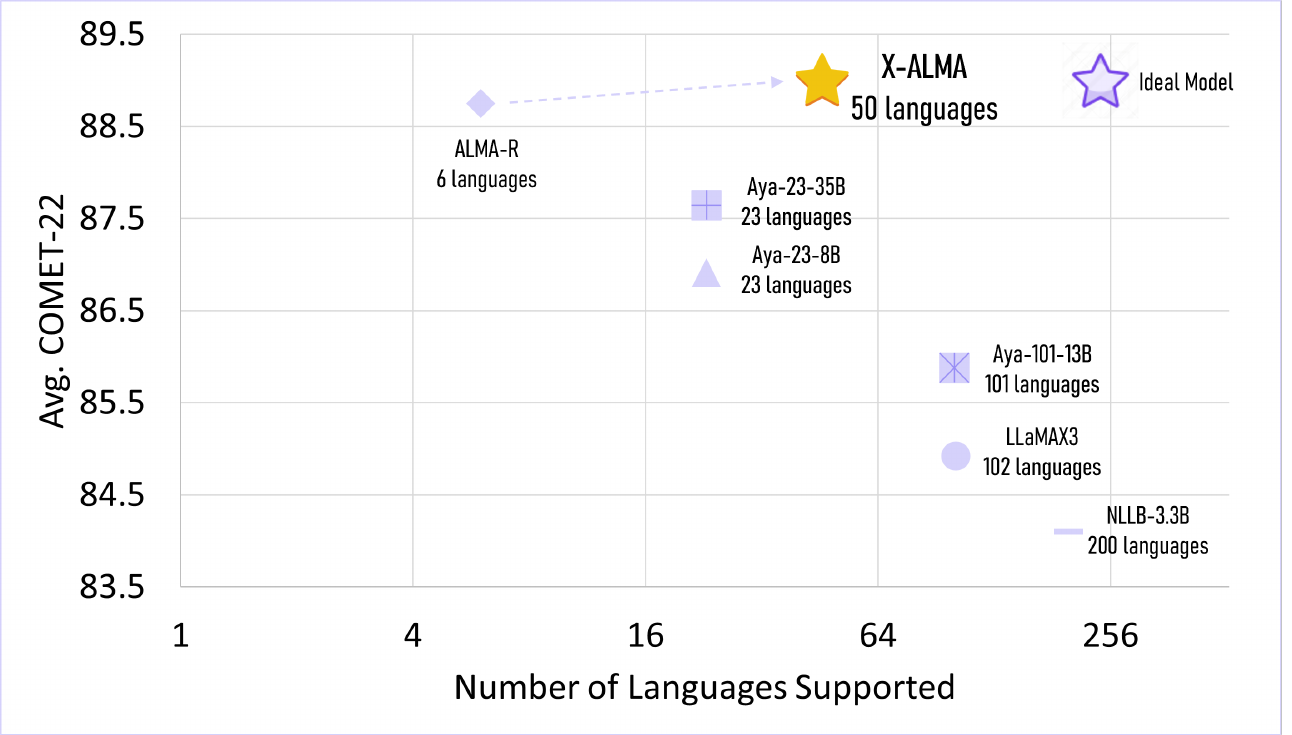}}
    \caption{
    Depiction of the general inverse trend  between the number of supported languages and average translation performance. While many state-of-the-art multilingual models claim to support hundreds of languages, the translation quality is not as high as in models trained on fewer languages, particularly for mid- and low-resource languages. This is reflected in the trend of decreasing average scores as more languages are supported. In contrast, we propose X-ALMA, which extends ALMA-R \citep{alma,alma-r} by supporting 44 additional diverse languages with even higher average performance, offering top performance across all supported languages, regardless of resource level.
    }
    \label{fig:all_models}
\end{figure}

% The above difficulty is intuitively understandable, as it is inherently challenging for mid- and low-resource languages to achieve the same level of performance as high-resource languages. 
To visualize these weaknesses, let us closely examine current models in the context of multilingual translation. We evaluate each model on the overlapping set of languages that are supported by the model and the 50 languages we focus on in this paper.\footnote{This is to depict a trend, and we acknowledge that scores are not directly comparable across languages.} As shown in Figure \ref{fig:all_models}, there is a clear trend: as the number of supported languages increases, the average translation performance decreases. This is intuitively understandable, as it is difficult for mid- and low-resource languages to reach the same level of performance as high-resource languages, thus lowering the overall average. For instance, ALMA-R \citep{alma-r} achieves the highest average translation performance across the 6 languages it supports, while NLLB-200 \citep{nllb} exhibits the lowest average performance on 50 languages, largely due to poorer results in low-resource languages. Although this comparison is not entirely fair due to the varying number of languages tested, it provides a general indication of above-mentioned weaknesses in multilingual models.\footnote{Here we evaluated these models using FLORES-200 \citep{nllb} test data and reporting the average COMET-22 \citep{comet22} across all languages, to or from English.}

Despite the ability of current multilingual models to support hundreds of languages, the hollow purple star `\inlinegraphics{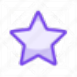}'
in the figure represents our ideal model, where the inclusion of more languages does not diminish the average performance. In this work, we introduce our multilingual translation model, \textbf{X-ALMA}, represented by the solid golden star `\inlinegraphics{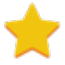}' in Figure \ref{fig:all_models}, which extends ALMA(-R) \citep{alma,alma-r} from 6 languages to 50 languages. ALMA-R is one of top-performing translation models built on LLMs, comparable to WMT winners and GPT-4-turbo. Despite the addition of 44 more languages, X-ALMA even achieves slightly higher average performance compared to ALMA-R.
%indicating that all added languages maintain high translation performance both into and from English. 
% To ensure a deep investigation and carefully address the needs of each language, we mainly focus on the multilingual machine translation task in this paper.

We summarize our main contributions as follows, including our model architecture design and training methodology.

\noindent\textbf{Plug-and-Play Architecture:} For capacity reasons, we design X-ALMA with several different  modules 
% to mitigate negative language interference\huda{"negative language interference" is not an established term AFAIK, i think you mean capactiy?},
with each module serving a group of similar languages. These modules can either be plugged into the base model individually for the inference of target languages---reducing the necessity of loading all expert parameters and saving memory---or all modules can be loaded together  in a  mixture-of-experts (MoE) way \citep{shazeer2017,gshard}.

\noindent\textbf{Effective Training Recipe:} The training regimen for X-ALMA consists of three pre-training stages and two post-training stages, each crucial for achieving optimal performance. Furthermore, in the final stage, we introduce \underline{\textbf{A}}daptive-\underline{\textbf{R}}ejection \underline{\textbf{P}}reference \underline{\textbf{O}}ptimization (\textbf{ARPO}), designed to maximize performance and address the `over-rejection' issue (detailed explanation in Section \ref{sec:adaptive_reject}) in translation preference learning, which current optimization methods struggle to resolve.
%The pre-training stages focus on acquiring general multilingual knowledge and alignment, and the first post-training stage is standard supervised fine-tuning (SFT). Here, we highlight the final stage, where we implement \textit{adaptive contrastive preference optimization (ARPO)} to mitigate the issue of 'over-rejection' in preference learning for translation. 'Over-rejection' occurs when the model generates translations in a style divergent from the preferred data, thereby diminishing or even compromising translation quality. This phenomenon arises because preferred and dis-preferred translations in preference data are typically very similar, differing by only a few words, which contrasts with other open-ended question-answering (QA) tasks. Conventional preference optimization struggles to handle such closely related pairs effectively.

\noindent\textbf{State-of-the-Art Performance and Data Release:} X-ALMA outperforms existing  open-source multilingual translation models \textit{across 50 diverse languages for every single direction} only training on publicly available data, as measured by COMET-22. 
% Given the limited availability of multilingual preference learning data,
To enable future work,
we also release the preference learning data for 50 languages and the model checkpoints.

% To have a deep investigation and carefully take care of each language, we mainly focus on the multilingual machine translation task with the goal of building a multilingual translation model that carefully addresses the needs of 50 diverse languages. 

% In this paper, we focus on the multilingual machine translation task with the goal of building a multilingual translation model that carefully addresses the needs of 50 diverse languages. Our aim is to ensure practical usability and high performance across both low-resource and high-resource languages.

\section{Background}
\subsection{Problem Definition}
We consider a decoder-only LLM, denoted as $\pi_{\theta}$, parameterized by $\theta$, for multilingual machine translation tasks. Let $\mathcal{D}$ represent the multilingual dataset, consisting of pairs of a source sentence $x$ and the corresponding perfect translation $y$, represented as $\mathcal{D}=\{x, y\}$. Given a prompt $\mathcal{I}$ that instructs the model to perform the translation, our goal is to maximize the log-likelihood of the multilingual parallel dataset $\mathcal{D}$: $\max_{\theta} \  \mathbb{E}_{(x,y) \sim \mathcal{D}} [\log \pi_\theta(y| x, \mathcal{I})]$.

% \begin{equation}
%       \max_{\theta} \  \mathbb{E}_{(x,y) \sim \mathcal{D}} [\log \pi_\theta(y| x, \mathcal{I})].
% \end{equation}

\subsection{Related Work}
\noindent\textbf{Multilingual Translation}
Massively Multilingual Translation models 
\citep{johnson-etal-2017-googles}, including open-source models such as PRISM \cite{thompson-post-2020-automatic,thompson-post-2020-paraphrase}, M2M-100 \citep{DBLP:journals/corr/abs-2010-11125}, and NLLB \citep{nllb} combine translation between many language pairs in a single encoder-decoder model.
% \footnote{M2M-100 specifically focused on non-English-centric translation, and NLLB continued that support. Our work does not.}
T5 \citep{10.5555/3455716.3455856} and mT5 \citep{xue-etal-2021-mt5} considered translation one of multitask learning. 

\noindent\textbf{LLM-Based Translation}
Initially, decoder-only LLMs struggled to match the performance of conventional encoder-decoder models for MT. For example, GPT-3.5 slightly under-performed the concurrent WMT winners \citep{gptmt}, and large open-source models like OPT-175B \citep{OPT} performed worse than the 1.3B parameter NLLB model \citep{nllb}, even on high-resource languages, as demonstrated by \citet{zhu-etal-2024-multilingual}. This lead to an increased interest in smaller LLMs, such as 7B or 13B models, because even smaller models like NLLB-1.3B showed strong translation capabilities. However, first generation LLM-based MT models such as TIM \citep{zeng2023tim}, SWIE \citep{swie}, and BayLing \citep{bayling} still lag behind encoder-decoder models in performance. The under performance of LLMs on translation lead to hybrid approaches combining LLMs with dedicated NMT models \citep{petrick-etal-2023-document,hoang-etal-2024-fly}. Recently, GPT-4 \citep{openai2023gpt4} has been reported to  achieve top performance in the WMT competition \citep{wmt23}, and smaller LLM-based models, like ALMA(-R) \citep{alma,alma-r} and Tower \citep{tower}, have demonstrated comparable performance to GPT-4 by employing their specialized training methods. However, the high performance of LLM-based translation models is limited to a small subset of languages.

\noindent\textbf{Massively Multilingual LLM}
The limited scope of languages in LLM-based MT models stems primarily from English-focused pre-training and the use of restricted vocabularies. However, this limitation has driven interest in expanding these models to support a broader range of languages. The simplest approach to extending current LLM-based MT models involves expanding the vocabulary and training on large amounts of parallel data across additional languages \citep{bigtranslate}, but this approach has been shown to degrade model performance. Aya-101 \citep{aya101} revisits the encoder-decoder architecture, building a multilingual model based on the largest MT5 \citep{mt5}, designed not only for translation but also for general multilingual QA. Similarly, LLaMAX \citep{llamax} extends LLaMA-2 and LLaMA-3 to over 100 languages. However, multilingual models often suffer from reduced performance on mid- and low-resource languages, which can also negatively impact high-resource language performance. To mitigate this, the decoder-only model Aya-23 \citep{aya23} focuses exclusively on 23 high-resource languages to maximize their performance and avoid the `curse of multilinguality'. While limiting the number of supported languages can indeed alleviate some challenges, it reverses the goal of building truly multilingual models and neglects the needs of mid- and low-resource languages. In this paper, we expand ALMA-R from 6 languages to 50, ensuring robust performance across all languages.

\section{Methods}
\subsection{Model Architecture}
\begin{figure}
    \centering
    \resizebox{0.5\linewidth}{!}{
    \includegraphics[width=7.5cm]{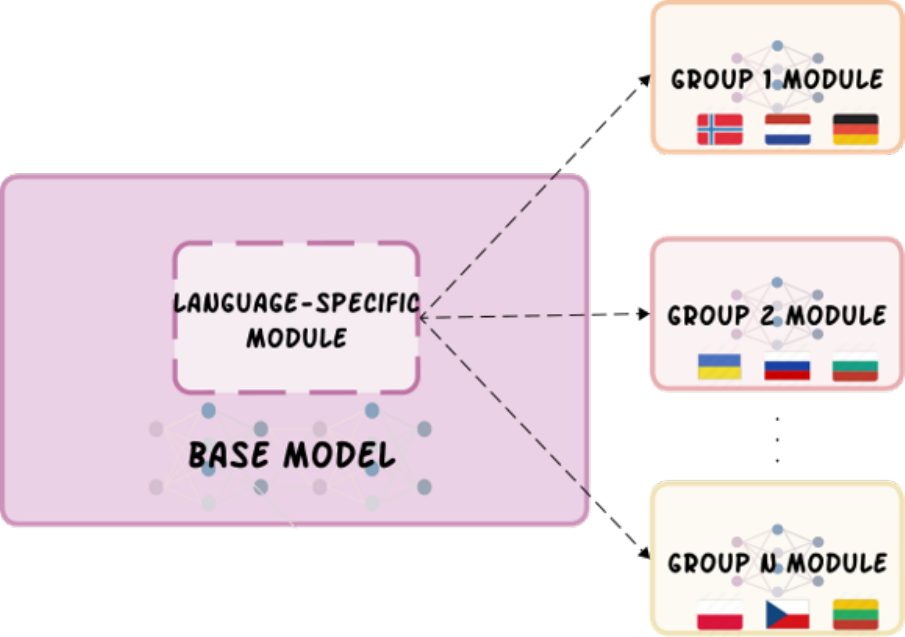}}
    \caption{
    High-level architecture design of the plug-and-play multilingual model. Each language group  is assigned a specific module that works alongside the base model. These language-specific modules handle inputs exclusively from their respective language groups, enabling the model to effectively adapt to different linguistic characteristics while leveraging the shared base model for comprehensive multilingual learning.
    }
    \label{fig:model_overview}
\end{figure}

Our model architecture consists of: (1) a dense base model, and (2) multiple language-specific (LS) modules. The core concept of LS modules is to prevent conflicts between languages during training, such as gradient conflicts \citep{wang2021gradient}. This design has similarities with the mixture-of-experts (MoE) approach \citep{shazeer2017,gshard}, but diverges by not using a neural-based gate to assign tokens to experts (LS modules). Instead, similar to \citet{xu-etal-2023-condensing}, the assignment is hard-gated, i.e., input data is assigned exclusively to the module designated for its language. Consequently, only the base model and the corresponding LS module are activated, depending on the input language. Languages are categorized into distinct groups, with each group sharing a common LS module. An overview of the model architecture is illustrated in Figure \ref{fig:model_overview}.

In detail, the base model architecture is built upon the LLaMA-2 architecture \citep{llama2}. Each LS module comprises low-rank adaptations (LoRAs) \citep{lora} integrated into all linear layers within the attention and multi-layer perceptron (MLP) layers. The total number of parameters for each LS module is approximately 15\% of the base model.

\noindent\textbf{Why This Design?}
While model architectures such as MoE activate only one expert per example, all experts must still reside in GPU memory during training and inference, necessitating high-end GPUs.
%and it is not practical to load only the activated parameters into the GPU.
Moreover, MoE has been reported for its parameter inefficiency in multilingual settings, e.g., hard-gated language assignment can achieve similar performance to MoE while using 4 times fewer parameters \citep{xu-etal-2023-condensing}. Compared to MoE, our design offers three distinct model-loading strategies for both training and inference: (1) selectively loading a single, on-demand LS module, which alleviates GPU memory constraints; (2) merging LS modules with the base model to generate a new LS LLM model that retains the same parameter count as the base model, facilitating subsequent use; and (3) loading both the base model and all LS modules as a larger, combined model, similar to the approach employed by MoE.

%\footnote{\href{https://huggingface.co/docs/peft/en/developer_guides/lora\#merge-lora-weights-into-the-base-model}{huggingface.co/merge-lora-weights}}

% \begin{wrapfigure}{r}{0.5\textwidth}
% \vskip -0.7in
%     \resizebox{1\linewidth}{!}{
%     \includegraphics[width=7.5cm]{figures/model_overview.pdf}}
%     \caption{
%     High-level architecture design of the plug-and-play multilingual model. Each language group (depicted as flags) is assigned a specific module that works alongside the base model. These language-specific modules handle inputs exclusively from their respective language groups, enabling the model to effectively adapt to different linguistic characteristics while leveraging the shared base model for comprehensive multilingual learning.
%     }
%     \label{fig:model_overview}
% \end{wrapfigure}
\subsection{Language Grouping}
\label{sec:language_group}
In this paper, we consider a total of 50 languages, encompassing 14 scripts and 18 language families, to capture the linguistic diversity. The languages are categorized into 8 groups based on two criteria: (1) each group should consist of languages that are as similar as possible, and (2) the number of languages in each group should be balanced. We opted not to use automated tools like Lang2Vec \citep{langvec} for grouping, as we found that manual grouping based on human linguistic knowledge yields more accurate classification in line with our criteria. The specific languages within each group are presented in Table \ref{tab:language_group} with their \texttt{ISO-639-1} code. Note that English (\texttt{en}) is included in all groups to ensure that each group can perform English-centric translation. More detailed information on these languages is provided in Appendix \ref{app:sec:lg_info}.

\begin{table}[t]
\caption{
Language grouping based on linguistic features and balanced number of languages.
}
\vskip 0.05in
\label{tab:language_group}
\centering

\resizebox{\linewidth}{!}{
\begin{tabular}{ccc}
\hline
Group ID & Linguistic Feature                     & Languages                            \\ \hline
1        & Germanic languages                     & \texttt{af}, \texttt{da}, \texttt{de}, \texttt{is}, \texttt{nl}, \texttt{no}, \texttt{sv}, (\texttt{en})     \\
2        & Romance Languages                      & \texttt{ca}, \texttt{es}, \texttt{gl}, \texttt{it}, \texttt{pt}, \texttt{ro}, (\texttt{en})         \\
3        & Eastern and Southern Slavic Languages  & \texttt{bg}, \texttt{mk}, \texttt{ru}, \texttt{sr}, \texttt{uk}, (\texttt{en})             \\
4        & Southeast Asian Languages              & \texttt{fr}, \texttt{id}, \texttt{mg}, \texttt{ms}, \texttt{th}, \texttt{vi}, (\texttt{en})         \\
5        & Central and Eastern European Languages & \texttt{cs}, \texttt{el}, \texttt{hu}, \texttt{lt}, \texttt{lv}, \texttt{pl}, (\texttt{en})         \\
6        & Eurasian Language Mix                  & \texttt{et}, \texttt{fi}, \texttt{ja}, \texttt{ka}, \texttt{ko}, \texttt{zh}, (\texttt{en})         \\
7        & Indo-Aryan Languages                   & \texttt{gu}, \texttt{hi}, \texttt{mr}, \texttt{ne}, \texttt{ur}, (\texttt{en})             \\
8        & Turkic and Semitic Languages           & \texttt{ar}, \texttt{az}, \texttt{fa}, \texttt{he}, \texttt{kk}, \texttt{ky}, \texttt{tr}, \texttt{uz}, (\texttt{en}) \\ \hline

\end{tabular}
}
\end{table}

\subsection{Training Recipe}
\label{sec:training_recipe}
We provide a comprehensive description of the training recipe for the X-ALMA model, including three pre-training stages and two post-training stages.  An overview of this training recipe is depicted in the workflow diagram in Appendix \ref{app:sec:training_recipe}. 
The specifics of each stage are elaborated upon as follows.

\noindent\textbf{Pre-Training Stage 1: Monolingual Fine-Tuning Base Model}
The first stage of pre-training is dedicated exclusively to the base model. During this phase, we fine-tune the base model using 20B monolingual tokens from all 50 languages, with a sampling ratio proportional to the size of the available monolingual data for each language, as suggested by \citet{alma}. This stage aims to facilitate the model's acquisition of fundamental knowledge across all languages.

\noindent\textbf{Pre-Training Stage 2: Monolingual Fine-Tuning Language-Specific Modules}
In all subsequent stages, the base model remains frozen, and the focus shifts to fine-tuning LS modules. During the second stage of pre-training, each LS module is fine-tuned with 10B monolingual tokens exclusively from the languages within its respective group. This stage is designed to enable each LS module to emphasize on learning  general knowledge across the specific languages.

\noindent\textbf{Pre-Training Stage 3: Pseudo-Monolingual Fine-tuning}
In this stage, we continue to fine-tune the LS modules using \textit{pseudo-monolingual data} from each module's language group. This pseudo-monolingual data is constructed from parallel sentences. While previous studies have indicated that simple instruction tuning with a large volume of parallel sentences for instruction tuning can degrade model performance \citep{alma,zhu-etal-2024-preference}, recent research demonstrates that utilizing parallel data in the pre-training stage can enhance multilingual alignment \citep{tower,kondo2024enhancing,llamax}. Similar to these approaches, we combine each available translation pair to create a new sentence in either a \textit{$<$source sentence$>$$<$target sentence$>$} or \textit{$<$target sentence$>$$<$source sentence$>$} manner, with the order of the source and target sentence in each pair determined randomly. We then concatenate all the combined translations to construct the pseudo-monolingual data. Each LS module is fine-tuned on 1.25B tokens.

\noindent\textbf{Post-Training Stage 1: Supervised Fine-tuning}
Building on the insights from prior research that small but high-quality multilingual datasets are sufficient to yield impressive performance \citep{maillard2023small,alma}, we supervised fine-tune (SFT) the model using a small, high-quality parallel dataset at this stage with the translation prompt suggested by \citet{alma}. This fine-tuning is performed using a simple causal language modeling (CLM) loss.

\noindent\textbf{Post-Training Stage 2: Preference Optimization}
We also introduce \textit{Adaptive Rejection Preference Optimization (ARPO)} to further enhance translation quality across all languages. ARPO is designed to address the `over-rejection' issue in MT preference learning, a challenge that other preference optimization methods struggle to manage effectively. We will elaborate on our motivations, methodology, and preference data construction in the following section.

\section{Adaptive-Rejection Preference Optimization}
\label{sec:adaptive_reject}
\subsection{Limitations in Current Preference Learning}
When constructing preference data for MT, it is essential that the dis-preferred translation is also of high quality to ensure meaningful model improvement \citep{alma-r}. This results in a scenario where the preferred and dis-preferred translations are often very similar, differing by only a few words, which is quite different from the preference data used in open-ended question-answering (QA) tasks (A detailed example is shown in Appendix \ref{app:cdf}). While many preference optimization methods have proven effective in various NLP tasks \citep{dpo,ipo,orpo,simpo}, we find that they are not well-suited for the MT task because they tend to reject the entire dis-preferred translation which is similar to the preferred one. This approach can inadvertently lead to the rejection of most tokens in the preferred translation as well, resulting in a phenomenon we term \textit{over-rejection}, where the writing style of the translation outputs is forced away from the preferred data distribution (further analysis and examples in Section \ref{sec:analysis_preference}).

Mathematically speaking, the preference optimization problem can be generally formulated given a dataset  $\mathcal{D} = \left\{ x^{(i)}, y_{w}^{(i)}, y_{l}^{(i)} \right\}_{i=1}^{N}$, where each data point consists of a prompt (source sentence) $x$, a preferred response (translation) $y_w$, and a dis-preferred response $y_l$, for a total of $N$ data points:
\begin{equation}
    \mathcal{L} =  \mathbb{E}_{(x,y_w,y_l) \sim \mathcal{D}} \Big[ f\Big( r_{\theta}(y_w | x) - r_{\theta}(y_l | x) \Big) \Big],
\label{eq:general_prefer}
\end{equation}
where $f(\cdot): \mathbb{R} \rightarrow \mathbb{R}$ is a general non-linear function. In many instances, such as in the DPO method \citep{dpo}, $f$ is the negative log-likelihood of the Bradley-Terry objective, i.e., $f(\cdot) = -\log \sigma(\cdot)$. Here, $r_{\theta}(y | x)$ represents the reward of $y$, calculated according to log-probability of the policy model parameterized by $\theta$. In the case of DPO, the reward function is defined as $r_{\theta}(y | x) = \beta\log(\pi_{\theta}(y | x)) - \beta\log(\pi_{\text{ref}}(y | x))$, where $\beta$ is a hyperparameter and $\pi_{\text{ref}}$ is the reference model.

As indicated by Equation \ref{eq:general_prefer}, when $y_w$ and $y_l$ are too similar, the difference between $r_{\theta}(y_w | x)$ and $r_{\theta}(y_l | x)$ tends to be small and even near 0. 
%If we take a closer look at practical implementations, the computation of $\log(\pi_{\theta}(y | x))$ used for deriving rewards is typically the sum of the log-likelihoods of all tokens in $y$.
Consequently, the near-zero difference between rewards causes the preference loss to a constant value, such as $f(0)=-\log \sigma(0)$ in the case of DPO. This makes it challenging for the optimization process to distinguish between the two options, hindering meaningful improvements.

The challenge becomes even more pronounced in a finite-data regime. While an infinite number of response pairs with small but precise preference differences could mitigate the optimization difficulties, translation preference data is often sparse and may contain noise (e.g., the AI-labeled preference data used by \citet{alma-r} contains only 2K samples per direction). Consequently, the model is prone to overfitting to these minor differences, which poses a significant empirical challenge and can lead to suboptimal learning outcomes, particularly when dealing with a large response (translation) space, as is the case with LLMs.

\subsection{Adaptive Rejection}
To mitigate the over-rejection, we introduce an adaptive penalty, denoted as $\tau_{\theta}$, which controls the strength of the dis-preferred term in the loss:
\begin{equation}
    \mathcal{L}_{\text{ARPO}} =  \mathbb{E}_{(x,y_w,y_l) \sim \mathcal{D}} \Big[ f\Big( r_{\theta}(y_w | x) - \tau_{\theta}(y_w, y_l)\cdot r_{\theta}(y_l | x) \Big) \Big].
\label{eq:general_penlaty}
\end{equation}
The value of $\tau_{\theta}$ is determined by the similarity between $y_w$ and $y_l$, ranging from $0$ to $1$:
\begin{equation}
    \tau_{\theta}(y_w, y_l) = \min(e^{\eta\cdot z_{\theta}(y_w, y_l)} - 1 ,1),
\label{eq:tau}
\end{equation}
where $\eta$ is a hyperparameter, and $z_{\theta}(y_w, y_l)$ is a function that quantifies the distance between the the preferred and dis-preferred responses by measuring absolute difference of their average log-likelihoods:
\begin{equation}
    z_{\theta}(y_w, y_l) = \text{abs}(\frac{\log(\pi_{\theta}(y_w | x))}{|y_w|} - \frac{\log(\pi_{\theta}(y_l | x))}{|y_l|}),    
\end{equation}
When $y_w$ and $y_l$ are very similar, the absolute difference between their averaged log-likelihoods is small, resulting in $\tau_{\theta}$ close to $0$, thereby reducing the impact of the dis-preferred term on the loss and mitigate rejection on this translation. Conversely, when the difference between $y_w$ and $y_l$ is large, $\tau_{\theta}$ close to $1$, turning the loss back to a standard preference optimization loss.

In the multilingual MT task, we start with contrastive preference optimization (CPO) \citep{alma-r}, which has proven to be one of the most effective optimization methods for translation.
\begin{equation}
\mathcal{L}_{\text{CPO}} =  -\mathbb{E}_{(x,y_w,y_l) \sim \mathcal{D}} \Big[ \underbrace{\log \sigma \Big( \beta \log \pi_{\theta}(y_w | x)  - \beta \log \pi_{\theta}(y_l | x) \Big)}_{\text{preference loss}} + \underbrace{\log \pi_\theta(y_w| x)}_{\text{BC loss}} \Big].
\label{eq:cpo}
\end{equation}
CPO consists of two components: preference loss and behavior cloning (BC) loss \citep{hejna2023contrastive}. The BC loss helps prevent the model from drifting too far from the original task. Then, we incorporate adaptive rejection into the preference term in CPO, resulting in a new loss function:
\begin{equation}
\mathcal{L}_{\text{ARPO}} =  -\mathbb{E}_{(x,y_w,y_l) \sim \mathcal{D}} \Big[ \log \sigma \Big( \beta \log \pi_{\theta}(y_w | x) - \tau_{\theta}(y_w, y_l) \cdot\beta \log \pi_{\theta}(y_l | x) \Big) + \log \pi_\theta(y_w| x) \Big].
\label{eq:acpo}
\end{equation}

\section{Experiments}
\label{sec:experiment}
\subsection{Data}
\label{sec:data}
% \noindent\textbf{Monolingual and Parallel Data} 
\paragraph{Monolingual and Parallel Data}
Following the introduction of 50 languages in Section \ref{sec:language_group}, we focus on 98 English-centric translation directions, both into and from English. We test on Flores-200 test data \citep{nllb} and WMT'23 \citep{wmt23}. For pre-training stages 1 and 2, we use monolingual data  from OSCAR \citep{OrtizSuarezSagotRomary2019}. In pre-training stage 3, we construct pseudo-monolingual data using NLLB \citep{schwenk-etal-2021-ccmatrix, heffernan-etal-2022-bitext, nllb}\footnote{\url{https://huggingface.co/datasets/allenai/nllb}} and OPUS \citep{tiedemann-2012-parallel,improvingmmt} parallel training data. Web crawled data (included in NLLB) has been shown to contain substantial mis-aligned and mis-translated segments \citep{khayrallah-koehn-2018-impact,10.1162/tacl_a_00447} and low-quality machine translated segments \citep{thompson-etal-2024-shocking}. Therefore,  in the SFT step---building on the insights from \citet{alma} that a small amount of \textit{high-quality} data can significantly enhance translation performance---we use the Flores-200 dev set and NTREX \citep{wmt19,federmann-etal-2022-ntrex} test data as our training data to ensure the quality. Given that both Flores-200 and NTREX are multi-way-parallel datasets (all languages share the same English source sentences),  we also incorporate the WMT'15-22 test data in training. 
The final data size in the SFT stage for each direction ranges from 3K to 7K, with an average of 4K per direction.

% \noindent\textbf{Preference Data Construction}
\paragraph{Preference Data Construction}
Given the scarcity of preference datasets for multilingual MT, we describe our approach to constructing preference data for 50 languages. Starting with the parallel data used in SFT, for each source sentence $x$, we generate a translation $y_\text{xalma}$ using X-ALMA that has been fine-tuned through SFT. Then, the reference translation $y_\text{ref}$ is designated as the preferred translation, and $y_\text{xalma}$ as the dis-preferred one, forming our initial preference dataset, denoted as $\mathcal{D}_1 = \{x, y_\text{ref}, y_\text{xalma}\}$.
Unlike \citet{alma-r}, we avoid the use of reference-free methods like XCOMET \citep{xcomet} for ranking translations in preference data construction to avoid potential bias, as the same metrics are used for evaluation.
% While \citet{alma-r} suggests that reference translations can sometimes be inferior to system-generated ones and employs reference-free methods like XCOMET \citep{xcomet} for ranking translations in preference data construction, we opted not to use this method to avoid potential metric bias, as the same metrics are used for evaluation.
As a result, $\mathcal{D}1$ might contain some noise due to the assumption that reference translations are always preferred. To reduce this noise, for high-resource languages, we also employ GPT-4o to produce revised translations $y_\text{gpt}$ conditioned on $(x, y_\text{xalma})$, drawing on studies that show post-editing by LLMs can improve translation quality \citep{ki2024guiding,feng2024ladder,raunak-etal-2023-leveraging}.
We show the prompts in Appendix \ref{app:sec:prompt}.
Thus, our second preference dataset is defined as $\mathcal{D}_2 = \{x, y_\text{gpt}, y_\text{xalma}\}$. We then concatenate the two datasets to form the final preference dataset, denoted as $\mathcal{D} = \mathcal{D}_1 \cup \mathcal{D}_2$.

\subsection{evaluation}
We report COMET-22 \citep{comet22} as our main metric as suggested by  \cite{freitag-etal-2023-results, freitag-etal-2024-llms}. In Appendix \ref{app:sec:full_results}, we also include XCOMET-XL (without references) \citep{xcomet} as recommended by \citet{alma-r}, and BLEU  \citep{papineni2002bleu,sacrebleu} for completeness.

\subsection{Training Setup}
We use ALMA-13B-Pretrain \citep{alma} as our backbone model, which is pre-trained on 6 languages and based on LLaMA-2 \citep{llama2}. Following \citet{alma}, we pre-train the backbone model with a batch size of 256, a warm-up ratio of 0.01, and sequences containing up to 512 tokens. In the post-training stage, the model is fine-tuned for many-to-many multilingual translation manner using 1 epoch with a batch size of 128, and other settings remain unchanged. For preference learning, we set $\eta$ as $1.5$ and $\beta$ as $0.1$ for all experiments.

\subsection{Baselines}
We use the strongest open-source massively multilingual translation models as our baselines, including NLLB-200 \citep{nllb}, Aya-101 \citep{aya101}, and LLaMAX3-Alpaca \citep{llamax}. Additionally, we compare our model's translation performance with Aya-23-8B and Aya-35B \citep{aya23} to demonstrate that increasing the number of supported languages does not compromise the performance of high-resource languages, effectively mitigating the curse of multilinguality. We also include LLaMA-3.1-8B-Instruct as a baseline to assess the performance of one of the latest strong LLMs in multilingual translation.

\subsection{Results}
\begin{table}[t]
\caption{
The overall results of Flores test data across each language group in \texttt{en}$\rightarrow$\texttt{xx}. Scores are reported using COMET-22. X-ALMA outperforms both massively multilingual models, such as Aya-101, and models focus specifically on high-resource languages, like Aya-23. `All' represents the average performance across all languages in the group, while `High' refers to the average performance for high-resource languages in the group. \textbf{Bold numbers} represent the highest scores.
}
\vskip 0.05in
\label{tab:main_en_xx}
\centering

\resizebox{1\linewidth}{!}{
\begin{tabular}{lcccccccccccccccc}
\hline
\multirow{2}{*}{Models} & \multicolumn{2}{c}{Group 1}   & \multicolumn{2}{c}{Group 2}   & \multicolumn{2}{c}{Group 3}   & \multicolumn{2}{c}{Group 4}   & \multicolumn{2}{c}{Group 5}   & \multicolumn{2}{c}{Group 6}   & \multicolumn{2}{c}{Group 7}   & \multicolumn{2}{c}{Group 8}   \\
\addlinespace[-0.5ex]
\cmidrule(lr){2-3} \cmidrule(lr){4-5} \cmidrule(lr){6-7}  \cmidrule(lr){8-9}  \cmidrule(lr){10-11} \cmidrule(lr){12-13} \cmidrule(lr){14-15}    \cmidrule(lr){16-17} 
\addlinespace[-0.5ex]
                        & All           & High          & All           & High          & All           & High          & All           & High          & All           & High          & All           & High          & All           & High          & All           & High          \\ \hline
LLaMA-3.1-8B-Instruct               & 80.8          & 79.8          & 83.7          & 84.2          & 79.1          & 69.2          & 76.3          & 85.8          & 79.0          & 81.2          & 71.1          & 71.8          & 70.1          & 69.6          & 78.3          & 84.4          \\
NLLB-3.3B               & 88.2          & 88.8          & 88.3          & 88.1          & 89.4          & 89.1          & 87.1          & 88.2          & 89.2          & 89.8          & 87.5          & 87.5          & 80.1          & 80.9          & 88.1          & 87.5          \\
LLaMAX3-Alpaca-8B                & 86.4          & 86.9          & 86.8          & 86.6          & 85.7          & 82.0          & 81.7          & 86.2          & 86.6          & 87.1          & 86.0          & 87.3          & 76.5          & 76.6          & 82.6          & 83.6          \\
Aya-101                 & 85.0          & 85.7          & 86.8          & 86.2          & 87.7          & 85.6          & 85.8          & 85.5          & 88.4          & 88.7          & 87.5          & 87.3          & 76.2          & 75.5          & 86.8          & 86.3          \\
Aya-23-8B               & 75.1          & 84.7          & 86.6          & 86.6          & 74.4          & 75.7          & 74.6          & 88.7          & 70.6          & 77.3          & 67.1          & 79.8          & 68.9          & 79.3          & 76.0          & 87.9          \\
Aya-23-35B              & 79.6          & 86.5          & 87.1          & 87.0          & 77.6          & 78.5          & 76.7          & 88.6          & 82.1          & 86.0          & 73.9          & 84.4          & 61.9          & 79.1          & 68.8          & 87.8          \\
\hdashline
X-ALMA (only SFT)       & 89.5          & 89.7          & 89.2          & 88.9          & 90.7          & 90.2          & 88.1          & 89.1          & 90.6          & 90.7          & 90.1          & 90.4          & 82.6          & 81.4          & 89.2          & 88.9          \\
X-ALMA                  & \textbf{89.6} & \textbf{89.9} & \textbf{89.4} & \textbf{89.0} & \textbf{90.9} & \textbf{90.5} & \textbf{88.6} & \textbf{89.5} & \textbf{91.0} & \textbf{91.1} & \textbf{90.6} & \textbf{90.8} & \textbf{83.2} & \textbf{81.9} & \textbf{89.4} & \textbf{89.2} \\ \hline
\end{tabular}
}
\end{table}

\begin{table}[t]
\caption{
The overall COMET-22 scores of  Flores test data across each language group in \texttt{xx}$\rightarrow$\texttt{en}. Similarly, X-ALMA outperforms all baselines.
}
\vskip 0.05in
\label{tab:main_xx_en}
\centering

\resizebox{1\linewidth}{!}{
\begin{tabular}{lcccccccccccccccc}
\hline
\multirow{2}{*}{Models} & \multicolumn{2}{c}{Group 1}   & \multicolumn{2}{c}{Group 2}   & \multicolumn{2}{c}{Group 3}   & \multicolumn{2}{c}{Group 4}   & \multicolumn{2}{c}{Group 5}   & \multicolumn{2}{c}{Group 6}   & \multicolumn{2}{c}{Group 7}   & \multicolumn{2}{c}{Group 8}   \\
\addlinespace[-0.5ex]
\cmidrule(lr){2-3} \cmidrule(lr){4-5} \cmidrule(lr){6-7}  \cmidrule(lr){8-9}  \cmidrule(lr){10-11} \cmidrule(lr){12-13} \cmidrule(lr){14-15}    \cmidrule(lr){16-17} 
\addlinespace[-0.5ex]
                        & All           & High          & All           & High          & All           & High          & All           & High          & All           & High          & All           & High          & All           & High          & All           & High          \\ \hline
LLaMA-3.1-8B-Instruct               & 68.8          & 77.6          & 70.9          & 76.9          & 51.2          & 53.8          & 65.6          & 76.4          & 54.8          & 60.2          & 58.8          & 66.9          & 47.6          & 53.7          & 53.7          & 67.5          \\
NLLB-3.3B               & 79.1          & 81.8          & 84.5          & 85.0          & 84.3          & 83.8          & 81.1          & 85.4          & 74.9          & 76.0          & 76.1          & 77.3          & 88.3          & 88.9          & 79.5          & 81.6          \\
LLaMAX3-Alpaca-8B              & 88.3          & 88.5          & 88.1          & 87.9          & 87.0          & 86.8          & 86.2          & 87.9          & 87.0          & 87.2          & 81.7          & 87.7          & 83.6          & 88.9          & 84.7          & 87.7          \\
Aya-101                 & 87.2          & 88.2          & 87.6          & 87.6          & 85.4          & 85.5          & 86.2          & 87.6          & 86.3          & 86.5          & 86.5          & 86.7          & 84.8          & 87.5          & 85.9          & 87.1          \\
Aya-23-8B               & 84.6          & 88.2          & 87.9          & 87.7          & 83.3          & 83.3          & 79.8          & 88.5          & 82.9          & 85.2          & 79.9          & 86.1          & 71.8          & 89.1          & 76.7          & 88.0          \\
Aya-23-35B              & 87.4          & 88.9          & 88.8          & 88.6          & 86.3          & 86.2          & 82.3          & 88.7          & 86.4          & 87.2          & 85.9          & 88.0          & 79.4          & 89.6          & 82.7          & 88.6          \\ \hdashline
X-ALMA (only SFT)       & 89.1          & 89.2          & 88.8          & 88.6          & 87.9          & 87.7          & 87.7          & 88.8          & 87.9          & 88.1          & 88.2          & 88.3          & 89.3          & 89.8          & 87.5          & 88.4          \\
X-ALMA                  & \textbf{89.4} & \textbf{89.5} & \textbf{89.2} & \textbf{89.0} & \textbf{88.1} & \textbf{87.8} & \textbf{88.0} & \textbf{88.9} & \textbf{88.2} & \textbf{88.4} & \textbf{88.7} & \textbf{88.8} & \textbf{89.6} & \textbf{90.1} & \textbf{88.0} & \textbf{88.9} \\ \hline
\end{tabular}
}
\end{table}

\begin{table}[t]
\caption{
Results on WMT'23 dataset reported using COMET-22.  The symbol $\rightarrow$ represents translations from English into the target language, while $\leftarrow$ indicates translations into English.
}
\vskip 0.05in
\label{tab:main_wmt23}
\centering

\resizebox{1\linewidth}{!}{
\begin{tabular}{lcccccccccccccc}
\hline
\multirow{2}{*}{Models} & \multicolumn{2}{c}{\texttt{de}}        & \multicolumn{2}{c}{\texttt{zh}}        & \multicolumn{2}{c}{\texttt{ja}}        & \multicolumn{2}{c}{\texttt{ru}}        & \multicolumn{2}{c}{\texttt{uk}}        & \multicolumn{2}{c}{\texttt{he}}        & \multicolumn{2}{c}{Avg.}    
  \\
                        & $\rightarrow$  & $\leftarrow$  & $\rightarrow$  & $\leftarrow$  & $\rightarrow$  & $\leftarrow$  & $\rightarrow$  & $\leftarrow$  & $\rightarrow$  & $\leftarrow$  & $\rightarrow$  & $\leftarrow$  & $\rightarrow$  & $\leftarrow$  \\ \hline
ALMA-R-13B               & 84.0           & 85.5            & 85.0           & 80.6            &  -           & -            & 85.5          & \textbf{83.3}            & -           & -            &  -           & -            & -           & -            \\ 
TowerInstruct-7B-v0.2               & 83.1           & 84.6            & 85.6           & 80.5            &  -           & -            & 85.3          & 83.1            & -           & -            &  -           & -            & -           & -            \\ 

\hdashline
NLLB-3.3B               & 79.7           & 66.6            & 79.6           & 67.8            & 81.6           & 65.8            & 83.8           & 76.7            & 82.8           & 79.0            & 83.6           & 79.9            & 81.8           & 72.6            \\
LlamaX3-8B              & 73.3           & 79.4            & 81.5           & 79.3            & 81.8           & 80.1            & 81.6           & 81.3            & 80.6           & 84.9            & 82.5           & 83.0            & 80.2           & 81.3            \\
Aya-101                 & 75.1           & 81.6            & 78.6           & 73.7            & 84.6           & 77.3            & 83.1           & 81.4            & 82.7           & 84.5            & 82.0           & 82.9            & 81.0           & 80.2            \\
Aya-23-8B               & 80.4           & 82.1            & 85.3           & 78.8            & 86.5           & 80.2            & 84.3           & 81.6            & 84.3           & 85.0            & 84.3           & 84.9            & 84.2           & 82.1            \\
Aya-23-35B              & 80.7           & 82.3            & 84.6           & 79.7            & 86.4           & 81.6            & 84.7           & 82.2            & 84.0           & 85.7            & 84.1           & 85.9            & 84.1           & 82.9            \\ \hdashline
X-ALMA (only SFT)       & 84.1           & 85.3            & 86.1           & 80.3            & 86.8           & 81.6            & 85.9           & 82.4            & 85.3           & 86.4            & 86.1           & 84.4            & 85.7           & 83.4            \\
X-ALMA                  & \textbf{84.4}  & \textbf{85.7}   & \textbf{86.7}  & \textbf{80.9}   & \textbf{87.5}  & \textbf{82.4}   & \textbf{86.3}  & \textbf{83.3}   & \textbf{85.5}  & \textbf{86.8}   & \textbf{86.2}  & \textbf{85.6}   & \textbf{86.1}  & \textbf{84.1}   \\ \hline
\end{tabular}
}
\end{table}

We present the average performance for each language group in both \texttt{en}$\rightarrow$\texttt{xx} and \texttt{xx}$\rightarrow$\texttt{en} directions on the Flores-200 test data in Tables \ref{tab:main_en_xx} and \ref{tab:main_xx_en}. The results for WMT'23 in both directions are provided in Table \ref{tab:main_wmt23}.  Detailed results for each translation direction can also be found in Appendix \ref{app:sec:full_results}.

\noindent\textbf{Compared with SoTA Multilingual Open Models:} 
General instruction-tuned LLaMA-3.1 significantly lags behind models specifically designed for translation, so we primarily focus on other models. 
 X-ALMA  outperforms other massively multilingual models such as NLLB-3.3B, LLaMAX3-Alpaca-8B, and Aya-101 on average across all language groups, both into and from English for 
both Flores-200 and WMT'23 test sets.  Furthermore, X-ALMA surpasses Aya-23-8B and Aya-23-35B---both of which are tailored for high-resource languages---on average across all high-resource languages in each group. In fact, as detailed in Appendix \ref{app:sec:full_results}, \textit{X-ALMA surpasses all baselines in all translation directions according to COMET-22 and outperforms in 97 out of 98 directions based on XCOMET-XL}, achieving top translation performance for all languages considered.

\noindent\textbf{Effectiveness of ARPO:}
ARPO delivers consistent improvements compared to SFT-only models in Flores-200 and WMT'23. Similarly, as shown in the full results in Appendix \ref{app:sec:full_results}, \textit{ARPO enhances performance in every translation direction as measured by COMET-22 and delivers improvements in 95 out of 98 directions according to XCOMET-XL.} We also compare the effectiveness of ARPO against other preference optimization methods in Section \ref{sec:analysis_preference}.

\section{Analysis}
All analyses are conducted on languages in Group 6, as it is the most challenging group to learn due to its mix of typologically  diverse Asian and European languages.
\subsection{Preference Optimization Comparison}
\label{sec:analysis_preference}
Here, we compare ARPO with other popular optimization methods, including DPO \citep{dpo}, KTO \citep{kto}, ORPO \citep{orpo}, SimPO \citep{simpo}, and the original CPO \citep{alma-r}. As indicated by CPO findings, directly applying preference learning to the MT task can harm the model, but adding a behavior cloning (BC) regularizer can stabilize training and improve the performance \citep{alma-r}. Following them, we also incorporate a BC regularizer into optimization methods that do not originally include it to provide a fair comparison. Table \ref{tab:po_compare} presents the comparison of preference optimization methods across all three metrics. As shown, ARPO clearly outperforms all baselines.

\begin{table}[t]
\caption{
Average performance comparison of various preference optimization methods for \texttt{en}$\rightarrow$\texttt{xx} and \texttt{xx}$\rightarrow$\texttt{en} on Group 6.
}
\vskip 0.05in
\label{tab:po_compare}
\centering

\resizebox{1\linewidth}{!}{
\begin{tabular}{lcccccc}
\hline
\multirow{2}{*}{Models} & \multicolumn{3}{c}{Avg. en$\rightarrow$xx}    & \multicolumn{3}{c}{Avg. xx$\rightarrow$en}    \\
                        & BLEU          & COMET-22      & XCOMET-XL     & BLEU          & COMET-22      & XCOMET-XL     \\ \hline
XALMA (only SFT)        & 26.5          & 90.1          & 80.3          & 32.1          & 88.2          & 77.4          \\ \hdashline
\phantom{  +} + DPO                   & 0.7           & 53.6          & 51.1          & 7.1           & 79.2          & 64.6          \\
\phantom{pholder}+ BC              & 23.5          & 90.2          & 80.0          & 27.8          & 87.5          & 77.4          \\ \hdashline
\phantom{  +} + SimPO                 & 0.0           & 16.7          & 1.5           & 0.0           & 16.4          & 8.9           \\
\phantom{pholder}+ BC            & 23.3          & 89.7          & 78.7          & 26.6          & 87.1          & 76.5          \\ \hdashline
\phantom{  +} + KTO                   & 22.1          & 89.8          & 79.2          & 26.4          & 87.1          & 76.5          \\
\phantom{pholder}+ BC              & 26.4          & 90.3          & 80.4          & 29.2          & 87.5          & 77.1          \\ \hdashline
\phantom{  +} + ORPO                  & 23.0          & 85.8          & 75.9          & 22.7          & 81.8          & 70.8          \\ \hdashline
\phantom{  +} + CPO                   & 22.2          & 90.2          & 79.9          & 26.5          & 87.8          & 77.0          \\ \hdashline
\phantom{  +} + ARPO (Final X-ALMA)     & \textbf{27.8} & \textbf{90.6} & \textbf{81.3} & \textbf{32.2} & \textbf{88.7} & \textbf{78.4} \\ \hline
\end{tabular}
}
\end{table}

\paragraph{Over-Rejection}
Over-rejection manifests itself under the significant shift in writing style away from the preferred data distribution. We observe a clear and big BLEU scores drop across all other preference optimization methods, indicating a decline in lexical matching. However, for certain methods, such as DPO + BC, KTO + BC, and CPO, both COMET scores do not decrease as drastically (and in some cases even improve slightly for \texttt{en}$\rightarrow$\texttt{xx}), suggesting that the models still produce accurate translations that maintain the same semantic meaning, but with a different writing style. Some translation examples generated by CPO are provided in Appendix \ref{app:sec:over_reject}. Unlike ARPO, methods such as CPO tend to produce a wider range of writing styles to convey the same meaning as the reference, many of which are accurate and non-detrimental. However, excessive shifts in writing style still can introduce translation errors that negatively impact overall quality. These small number of errors, concealed within a lot of stylistic deviations, are where over-rejection occurs. As hypothesized in Section \ref{sec:adaptive_reject}, the significant style shift is caused by the model rejecting dis-preferred translations that are similar to preferred ones, leading to an excessive rejection of certain writing styles from the preferred data. However, ARPO addresses this issue by \textit{constraining the stylistic variation within a more controlled range}, thereby mitigating errors caused by over-rejection.

For other methods such as naive DPO and SimPO, which even though work well in other NLP tasks, the over-rejection severely impairs the model's ability to generate meaningful translations. The introduction of ARPO significantly mitigate the over-rejection issue (stable BLEU scores) and maximize the translation qualities (the highest scores in two COMET metrics).

\subsection{Ablation Study}
\begin{figure}[ht]
    \centering
    \resizebox{1\linewidth}{!}{
    \includegraphics[width=7.5cm]{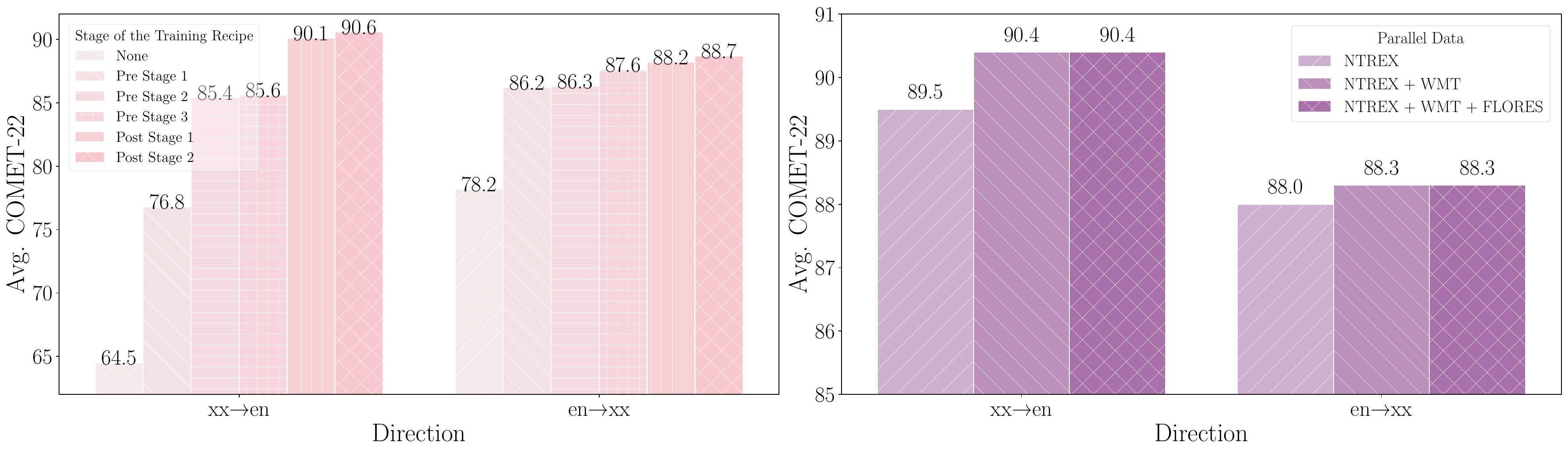}}
    \caption{
    \textbf{Left:} ablation study on each stage of the training recipe, demonstrating that adding each stage leads to consistent performance improvements. \textbf{Right:} ablation study on the impact of parallel data composition during the SFT stage. Adding WMT data to NTREX significantly enhances model performance, while adding Flores-200 data provides no noticeable improvement.
    }
    \label{fig:ablation}
\end{figure}

\paragraph{Training Recipe}
We investigate the impact of each step in the training recipe on model performance. The average results for Group 6, both into and from English, are presented in the left part of Figure \ref{fig:ablation}. The results show a clear trend of consistent performance improvement with each step in the training process. Note that `None' is the initial checkpoint in our recipe, ALMA-13B-Pretrain.

\paragraph{Parallel Data for SFT}
For SFT, we use high-quality parallel data from three sources: NTREX, WMT, and the Flores-200 dev set. Here, we investigate how combining parallel datasets affects performance \textit{during the SFT stage}. As shown on the right of Figure \ref{fig:ablation}, using only NTREX data already achieves impressive average translation performance for Group 6. Adding high-quality WMT data further boosts average performance, particularly for translations into English data. We hypothesize that this improvement stems from the increased diversity of English data, which mitigates overfitting to the NTREX English domain—a known issue with multi-way-parallel data, as observed by \citet{aharoni-etal-2019-massively}. Conversely, incorporating more Flores-200 dev data (also multi-way-parallel) into training does not result in significant gains, also suggesting that the strong translation performance is not driven by in-domain Flores-200 data.

\section{Conclusion}
We tackled the challenge of achieving high translation quality while scaling to a large number of languages, a limitation seen in many state-of-the-art multilingual models. We have introduced X-ALMA, an LLM-based multilingual translation system that prioritizes translation quality across all supported 50 languages, regardless of resource level. X-ALMA surpasses SoTA open models such as Aya-101 and Aya-23 in all translation directions on the FLORES-200 and WMT'23 test datasets, as measured by COMET-22. X-ALMA is built on a plug-and-play architecture with language-specific modules, complemented by a carefully designed training recipe. In particular, the final stage of the recipe, ARPO, achieves further performance gains and outperforms existing preference optimization methods in translation tasks, while successfully mitigating the over-rejection issue.


\section*{Acknowledgements}
We thank anonymous reviewers for their insightful feedback. We express our profound appreciation to HyoJung Han, Jack Zhang, Tianjian Li, Thamme Gowda, Tom Kocmi, Young Jin Kim, Hany Hassan Awadalla, Marcin Junczys-Dowmunt, Vikas Raunak, Matt Post, Anoop Kunchukuttan, Roman Grundkiewicz, Arul Menezes, and Vishal Chowdhary for their engaging and valuable discussions that greatly enriched our work.

\bibliography{iclr2025_conference}

\begin{thebibliography}{70}
\providecommand{\natexlab}[1]{#1}
\providecommand{\url}[1]{\texttt{#1}}
\expandafter\ifx\csname urlstyle\endcsname\relax
  \providecommand{\doi}[1]{doi: #1}\else
  \providecommand{\doi}{doi: \begingroup \urlstyle{rm}\Url}\fi

\bibitem[Aharoni et~al.(2019)Aharoni, Johnson, and Firat]{aharoni-etal-2019-massively}
Roee Aharoni, Melvin Johnson, and Orhan Firat.
\newblock Massively multilingual neural machine translation.
\newblock In \emph{Proceedings of the 2019 Conference of the North {A}merican Chapter of the Association for Computational Linguistics: Human Language Technologies, Volume 1 (Long and Short Papers)}, pp.\  3874--3884, Minneapolis, Minnesota, June 2019. Association for Computational Linguistics.
\newblock \doi{10.18653/v1/N19-1388}.
\newblock URL \url{https://aclanthology.org/N19-1388}.

\bibitem[Alves et~al.(2024)Alves, Pombal, Guerreiro, Martins, Alves, Farajian, Peters, Rei, Fernandes, Agrawal, et~al.]{tower}
Duarte~M Alves, Jos{\'e} Pombal, Nuno~M Guerreiro, Pedro~H Martins, Jo{\~a}o Alves, Amin Farajian, Ben Peters, Ricardo Rei, Patrick Fernandes, Sweta Agrawal, et~al.
\newblock Tower: An open multilingual large language model for translation-related tasks.
\newblock \emph{arXiv preprint arXiv:2402.17733}, 2024.

\bibitem[Aryabumi et~al.(2024)Aryabumi, Dang, Talupuru, Dash, Cairuz, Lin, Venkitesh, Smith, Marchisio, Ruder, et~al.]{aya23}
Viraat Aryabumi, John Dang, Dwarak Talupuru, Saurabh Dash, David Cairuz, Hangyu Lin, Bharat Venkitesh, Madeline Smith, Kelly Marchisio, Sebastian Ruder, et~al.
\newblock Aya 23: Open weight releases to further multilingual progress.
\newblock \emph{arXiv preprint arXiv:2405.15032}, 2024.

\bibitem[Azar et~al.(2024)Azar, Guo, Piot, Munos, Rowland, Valko, and Calandriello]{ipo}
Mohammad~Gheshlaghi Azar, Zhaohan~Daniel Guo, Bilal Piot, Remi Munos, Mark Rowland, Michal Valko, and Daniele Calandriello.
\newblock A general theoretical paradigm to understand learning from human preferences.
\newblock In \emph{International Conference on Artificial Intelligence and Statistics}, pp.\  4447--4455. PMLR, 2024.

\bibitem[Barrault et~al.(2019)Barrault, Bojar, Costa-juss{\`a}, Federmann, Fishel, Graham, Haddow, Huck, Koehn, Malmasi, Monz, M{\"u}ller, Pal, Post, and Zampieri]{wmt19}
Lo{\"\i}c Barrault, Ond{\v{r}}ej Bojar, Marta~R. Costa-juss{\`a}, Christian Federmann, Mark Fishel, Yvette Graham, Barry Haddow, Matthias Huck, Philipp Koehn, Shervin Malmasi, Christof Monz, Mathias M{\"u}ller, Santanu Pal, Matt Post, and Marcos Zampieri.
\newblock Findings of the 2019 conference on machine translation ({WMT}19).
\newblock In Ond{\v{r}}ej Bojar, Rajen Chatterjee, Christian Federmann, Mark Fishel, Yvette Graham, Barry Haddow, Matthias Huck, Antonio~Jimeno Yepes, Philipp Koehn, Andr{\'e} Martins, Christof Monz, Matteo Negri, Aur{\'e}lie N{\'e}v{\'e}ol, Mariana Neves, Matt Post, Marco Turchi, and Karin Verspoor (eds.), \emph{Proceedings of the Fourth Conference on Machine Translation (Volume 2: Shared Task Papers, Day 1)}, pp.\  1--61, Florence, Italy, August 2019. Association for Computational Linguistics.
\newblock \doi{10.18653/v1/W19-5301}.
\newblock URL \url{https://aclanthology.org/W19-5301}.

\bibitem[Brown et~al.(2020)Brown, Mann, Ryder, Subbiah, Kaplan, Dhariwal, Neelakantan, Shyam, Sastry, Askell, et~al.]{gpt3_few_shot}
Tom Brown, Benjamin Mann, Nick Ryder, Melanie Subbiah, Jared~D Kaplan, Prafulla Dhariwal, Arvind Neelakantan, Pranav Shyam, Girish Sastry, Amanda Askell, et~al.
\newblock Language models are few-shot learners.
\newblock \emph{Advances in neural information processing systems}, 33:\penalty0 1877--1901, 2020.

\bibitem[Chen et~al.(2023)Chen, Liu, Meng, Chen, Xu, and Zhou]{swie}
Yijie Chen, Yijin Liu, Fandong Meng, Yufeng Chen, Jinan Xu, and Jie Zhou.
\newblock Improving translation faithfulness of large language models via augmenting instructions.
\newblock \emph{arXiv preprint arXiv:2308.12674}, 2023.

\bibitem[Conneau et~al.(2020)Conneau, Khandelwal, Goyal, Chaudhary, Wenzek, Guzm{\'a}n, Grave, Ott, Zettlemoyer, and Stoyanov]{xlmr}
Alexis Conneau, Kartikay Khandelwal, Naman Goyal, Vishrav Chaudhary, Guillaume Wenzek, Francisco Guzm{\'a}n, Edouard Grave, Myle Ott, Luke Zettlemoyer, and Veselin Stoyanov.
\newblock Unsupervised cross-lingual representation learning at scale.
\newblock In \emph{Proceedings of the 58th Annual Meeting of the Association for Computational Linguistics}, pp.\  8440--8451, Online, July 2020. Association for Computational Linguistics.
\newblock \doi{10.18653/v1/2020.acl-main.747}.
\newblock URL \url{https://aclanthology.org/2020.acl-main.747}.

\bibitem[Dubey et~al.(2024)Dubey, Jauhri, Pandey, Kadian, Al-Dahle, Letman, Mathur, Schelten, Yang, Fan, et~al.]{llama3}
Abhimanyu Dubey, Abhinav Jauhri, Abhinav Pandey, Abhishek Kadian, Ahmad Al-Dahle, Aiesha Letman, Akhil Mathur, Alan Schelten, Amy Yang, Angela Fan, et~al.
\newblock The llama 3 herd of models.
\newblock \emph{arXiv preprint arXiv:2407.21783}, 2024.

\bibitem[Ethayarajh et~al.(2024)Ethayarajh, Xu, Muennighoff, Jurafsky, and Kiela]{kto}
Kawin Ethayarajh, Winnie Xu, Niklas Muennighoff, Dan Jurafsky, and Douwe Kiela.
\newblock Kto: Model alignment as prospect theoretic optimization.
\newblock \emph{arXiv preprint arXiv:2402.01306}, 2024.

\bibitem[Fan et~al.(2020)Fan, Bhosale, Schwenk, Ma, El{-}Kishky, Goyal, Baines, Celebi, Wenzek, Chaudhary, Goyal, Birch, Liptchinsky, Edunov, Grave, Auli, and Joulin]{DBLP:journals/corr/abs-2010-11125}
Angela Fan, Shruti Bhosale, Holger Schwenk, Zhiyi Ma, Ahmed El{-}Kishky, Siddharth Goyal, Mandeep Baines, Onur Celebi, Guillaume Wenzek, Vishrav Chaudhary, Naman Goyal, Tom Birch, Vitaliy Liptchinsky, Sergey Edunov, Edouard Grave, Michael Auli, and Armand Joulin.
\newblock Beyond english-centric multilingual machine translation.
\newblock \emph{CoRR}, abs/2010.11125, 2020.
\newblock URL \url{https://arxiv.org/abs/2010.11125}.

\bibitem[Federmann et~al.(2022)Federmann, Kocmi, and Xin]{federmann-etal-2022-ntrex}
Christian Federmann, Tom Kocmi, and Ying Xin.
\newblock {NTREX}-128 {--} news test references for {MT} evaluation of 128 languages.
\newblock In \emph{Proceedings of the First Workshop on Scaling Up Multilingual Evaluation}, pp.\  21--24, Online, nov 2022. Association for Computational Linguistics.
\newblock URL \url{https://aclanthology.org/2022.sumeval-1.4}.

\bibitem[Feng et~al.(2024)Feng, Chen, Zhang, Meng, and Liu]{feng2024ladder}
Zhaopeng Feng, Ruizhe Chen, Yan Zhang, Zijie Meng, and Zuozhu Liu.
\newblock Ladder: A model-agnostic framework boosting llm-based machine translation to the next level.
\newblock \emph{arXiv preprint arXiv:2406.15741}, 2024.

\bibitem[Freitag et~al.(2023)Freitag, Mathur, Lo, Avramidis, Rei, Thompson, Kocmi, Blain, Deutsch, Stewart, Zerva, Castilho, Lavie, and Foster]{freitag-etal-2023-results}
Markus Freitag, Nitika Mathur, Chi-kiu Lo, Eleftherios Avramidis, Ricardo Rei, Brian Thompson, Tom Kocmi, Frederic Blain, Daniel Deutsch, Craig Stewart, Chrysoula Zerva, Sheila Castilho, Alon Lavie, and George Foster.
\newblock Results of {WMT}23 metrics shared task: Metrics might be guilty but references are not innocent.
\newblock In Philipp Koehn, Barry Haddow, Tom Kocmi, and Christof Monz (eds.), \emph{Proceedings of the Eighth Conference on Machine Translation}, pp.\  578--628, Singapore, December 2023. Association for Computational Linguistics.
\newblock \doi{10.18653/v1/2023.wmt-1.51}.
\newblock URL \url{https://aclanthology.org/2023.wmt-1.51/}.

\bibitem[Freitag et~al.(2024)Freitag, Mathur, Deutsch, Lo, Avramidis, Rei, Thompson, Blain, Kocmi, Wang, Adelani, Buchicchio, Zerva, and Lavie]{freitag-etal-2024-llms}
Markus Freitag, Nitika Mathur, Daniel Deutsch, Chi-Kiu Lo, Eleftherios Avramidis, Ricardo Rei, Brian Thompson, Frederic Blain, Tom Kocmi, Jiayi Wang, David~Ifeoluwa Adelani, Marianna Buchicchio, Chrysoula Zerva, and Alon Lavie.
\newblock Are {LLM}s breaking {MT} metrics? results of the {WMT}24 metrics shared task.
\newblock In Barry Haddow, Tom Kocmi, Philipp Koehn, and Christof Monz (eds.), \emph{Proceedings of the Ninth Conference on Machine Translation}, pp.\  47--81, Miami, Florida, USA, November 2024. Association for Computational Linguistics.
\newblock \doi{10.18653/v1/2024.wmt-1.2}.
\newblock URL \url{https://aclanthology.org/2024.wmt-1.2/}.

\bibitem[Guerreiro et~al.(2023)Guerreiro, Rei, van Stigt, Coheur, Colombo, and Martins]{xcomet}
Nuno~M Guerreiro, Ricardo Rei, Daan van Stigt, Luisa Coheur, Pierre Colombo, and Andr{\'e}~FT Martins.
\newblock xcomet: Transparent machine translation evaluation through fine-grained error detection.
\newblock \emph{arXiv preprint arXiv:2310.10482}, 2023.

\bibitem[Heffernan et~al.(2022)Heffernan, {\c{C}}elebi, and Schwenk]{heffernan-etal-2022-bitext}
Kevin Heffernan, Onur {\c{C}}elebi, and Holger Schwenk.
\newblock Bitext mining using distilled sentence representations for low-resource languages.
\newblock In Yoav Goldberg, Zornitsa Kozareva, and Yue Zhang (eds.), \emph{Findings of the Association for Computational Linguistics: EMNLP 2022}, pp.\  2101--2112, Abu Dhabi, United Arab Emirates, December 2022. Association for Computational Linguistics.
\newblock \doi{10.18653/v1/2022.findings-emnlp.154}.
\newblock URL \url{https://aclanthology.org/2022.findings-emnlp.154}.

\bibitem[Hejna et~al.(2023)Hejna, Rafailov, Sikchi, Finn, Niekum, Knox, and Sadigh]{hejna2023contrastive}
Joey Hejna, Rafael Rafailov, Harshit Sikchi, Chelsea Finn, Scott Niekum, W~Bradley Knox, and Dorsa Sadigh.
\newblock Contrastive prefence learning: Learning from human feedback without rl.
\newblock \emph{arXiv preprint arXiv:2310.13639}, 2023.

\bibitem[Hendy et~al.(2023)Hendy, Abdelrehim, Sharaf, Raunak, Gabr, Matsushita, Kim, Afify, and Awadalla]{gptmt}
Amr Hendy, Mohamed Abdelrehim, Amr Sharaf, Vikas Raunak, Mohamed Gabr, Hitokazu Matsushita, Young~Jin Kim, Mohamed Afify, and Hany~Hassan Awadalla.
\newblock How good are gpt models at machine translation? a comprehensive evaluation.
\newblock \emph{arXiv preprint arXiv:2302.09210}, 2023.

\bibitem[Hoang et~al.(2024)Hoang, Khayrallah, and Junczys-Dowmunt]{hoang-etal-2024-fly}
Hieu Hoang, Huda Khayrallah, and Marcin Junczys-Dowmunt.
\newblock On-the-fly fusion of large language models and machine translation.
\newblock In Kevin Duh, Helena Gomez, and Steven Bethard (eds.), \emph{Findings of the Association for Computational Linguistics: NAACL 2024}, pp.\  520--532, Mexico City, Mexico, June 2024. Association for Computational Linguistics.
\newblock \doi{10.18653/v1/2024.findings-naacl.35}.
\newblock URL \url{https://aclanthology.org/2024.findings-naacl.35/}.

\bibitem[Hong et~al.(2024)Hong, Lee, and Thorne]{orpo}
Jiwoo Hong, Noah Lee, and James Thorne.
\newblock Orpo: Monolithic preference optimization without reference model.
\newblock \emph{arXiv preprint arXiv:2403.07691}, 2\penalty0 (4):\penalty0 5, 2024.

\bibitem[Hu et~al.(2021)Hu, Wallis, Allen-Zhu, Li, Wang, Wang, Chen, et~al.]{lora}
Edward~J Hu, Phillip Wallis, Zeyuan Allen-Zhu, Yuanzhi Li, Shean Wang, Lu~Wang, Weizhu Chen, et~al.
\newblock Lora: Low-rank adaptation of large language models.
\newblock In \emph{International Conference on Learning Representations}, 2021.

\bibitem[Jiang et~al.(2023)Jiang, Sablayrolles, Mensch, Bamford, Chaplot, Casas, Bressand, Lengyel, Lample, Saulnier, et~al.]{mistral}
Albert~Q Jiang, Alexandre Sablayrolles, Arthur Mensch, Chris Bamford, Devendra~Singh Chaplot, Diego de~las Casas, Florian Bressand, Gianna Lengyel, Guillaume Lample, Lucile Saulnier, et~al.
\newblock Mistral 7b.
\newblock \emph{arXiv preprint arXiv:2310.06825}, 2023.

\bibitem[Johnson et~al.(2017)Johnson, Schuster, Le, Krikun, Wu, Chen, Thorat, Vi{\'e}gas, Wattenberg, Corrado, Hughes, and Dean]{johnson-etal-2017-googles}
Melvin Johnson, Mike Schuster, Quoc~V. Le, Maxim Krikun, Yonghui Wu, Zhifeng Chen, Nikhil Thorat, Fernanda Vi{\'e}gas, Martin Wattenberg, Greg Corrado, Macduff Hughes, and Jeffrey Dean.
\newblock {G}oogle`s multilingual neural machine translation system: Enabling zero-shot translation.
\newblock \emph{Transactions of the Association for Computational Linguistics}, 5:\penalty0 339--351, 2017.
\newblock \doi{10.1162/tacl_a_00065}.
\newblock URL \url{https://aclanthology.org/Q17-1024/}.

\bibitem[Khayrallah \& Koehn(2018)Khayrallah and Koehn]{khayrallah-koehn-2018-impact}
Huda Khayrallah and Philipp Koehn.
\newblock On the impact of various types of noise on neural machine translation.
\newblock In Alexandra Birch, Andrew Finch, Thang Luong, Graham Neubig, and Yusuke Oda (eds.), \emph{Proceedings of the 2nd Workshop on Neural Machine Translation and Generation}, pp.\  74--83, Melbourne, Australia, July 2018. Association for Computational Linguistics.
\newblock \doi{10.18653/v1/W18-2709}.
\newblock URL \url{https://aclanthology.org/W18-2709/}.

\bibitem[Ki \& Carpuat(2024)Ki and Carpuat]{ki2024guiding}
Dayeon Ki and Marine Carpuat.
\newblock Guiding large language models to post-edit machine translation with error annotations.
\newblock \emph{arXiv preprint arXiv:2404.07851}, 2024.

\bibitem[Kocmi et~al.(2023)Kocmi, Avramidis, Bawden, Bojar, Dvorkovich, Federmann, Fishel, Freitag, Gowda, Grundkiewicz, Haddow, Koehn, Marie, Monz, Morishita, Murray, Nagata, Nakazawa, Popel, Popovi{\'c}, and Shmatova]{wmt23}
Tom Kocmi, Eleftherios Avramidis, Rachel Bawden, Ond{\v{r}}ej Bojar, Anton Dvorkovich, Christian Federmann, Mark Fishel, Markus Freitag, Thamme Gowda, Roman Grundkiewicz, Barry Haddow, Philipp Koehn, Benjamin Marie, Christof Monz, Makoto Morishita, Kenton Murray, Makoto Nagata, Toshiaki Nakazawa, Martin Popel, Maja Popovi{\'c}, and Mariya Shmatova.
\newblock Findings of the 2023 conference on machine translation ({WMT}23): {LLM}s are here but not quite there yet.
\newblock In Philipp Koehn, Barry Haddow, Tom Kocmi, and Christof Monz (eds.), \emph{Proceedings of the Eighth Conference on Machine Translation}, pp.\  1--42, Singapore, December 2023. Association for Computational Linguistics.
\newblock URL \url{https://aclanthology.org/2023.wmt-1.1}.

\bibitem[Kondo et~al.(2024)Kondo, Utsuro, and Nagata]{kondo2024enhancing}
Minato Kondo, Takehito Utsuro, and Masaaki Nagata.
\newblock Enhancing translation accuracy of large language models through continual pre-training on parallel data.
\newblock \emph{arXiv preprint arXiv:2407.03145}, 2024.

\bibitem[Kreutzer et~al.(2022)Kreutzer, Caswell, Wang, Wahab, van Esch, Ulzii-Orshikh, Tapo, Subramani, Sokolov, Sikasote, Setyawan, Sarin, Samb, Sagot, Rivera, Rios, Papadimitriou, Osei, Suarez, Orife, Ogue\~ji, Rubungo, Nguyen, Müller, Müller, Muhammad, Muhammad, Mnyakeni, Mirzakhalov, Matangir\~a, Leong, Lawson, Kudugunta, Jernite, Jenny, Firat, Dossou, Dlamini, de~Silva, Çabuk Ballı\, Biderman, Battisti, Baruwa, Bapna, Baljekar, Azime, Awokoya, Ataman, Ahia, Ahia, Agrawal, and Adeyemi]{10.1162/tacl_a_00447}
Julia Kreutzer, Isaac Caswell, Lisa Wang, Ahsan Wahab, Daan van Esch, Nasanbayar Ulzii-Orshikh, Allahsera Tapo, Nishant Subramani, Artem Sokolov, Clayton\~e Sikasote, Monang Setyawan, Supheakmungkol Sarin, Sokhar Samb, Benoît Sagot, Clara Rivera, Annette Rios, Isabel Papadimitriou, Salomey Osei, Pedro~Ortiz Suarez, Iroro Orife, Kelechi Ogue\~ji, Andre~Niyongabo Rubungo, Toan~Q. Nguyen, Mathias Müller, André Müller, Shamsuddeen~Hassan Muhammad, Nanda Muhammad, Ayanda Mnyakeni, Jamshidbek Mirzakhalov, Tapiwanashe Matangir\~a, Colin Leong, Nze Lawson, Sneha Kudugunta, Yacine Jernite, Mathias Jenny, Orhan Firat, Bonaventure F.~P. Dossou, Sakhile Dlamini, Nisansa de~Silva, Sakine Çabuk Ballı\, Stella Biderman, Alessia Battisti, Ahmed Baruwa, Ankur Bapna, Pallavi Baljekar, Israel~Abebe Azime, Ayodele Awokoya, Duygu Ataman, Orevaoghene Ahia, Oghenefeg\~o Ahia, Sweta Agrawal, and Mofetoluwa Adeyemi.
\newblock Quality at a glance: An audit of web-crawled multilingual datasets.
\newblock \emph{Transactions of the Association for Computational Linguistics}, 10:\penalty0 50--72, 01 2022.
\newblock ISSN 2307-387X.
\newblock \doi{10.1162/tacl_a_00447}.
\newblock URL \url{https://doi.org/10.1162/tacl\_a\_00447}.

\bibitem[Lepikhin et~al.(2021)Lepikhin, Lee, Xu, Chen, Firat, Huang, Krikun, Shazeer, and Chen]{gshard}
Dmitry Lepikhin, HyoukJoong Lee, Yuanzhong Xu, Dehao Chen, Orhan Firat, Yanping Huang, Maxim Krikun, Noam Shazeer, and Zhifeng Chen.
\newblock {\{}GS{\}}hard: Scaling giant models with conditional computation and automatic sharding.
\newblock In \emph{International Conference on Learning Representations}, 2021.
\newblock URL \url{https://openreview.net/forum?id=qrwe7XHTmYb}.

\bibitem[Littell et~al.(2017)Littell, Mortensen, Lin, Kairis, Turner, and Levin]{langvec}
Patrick Littell, David~R. Mortensen, Ke~Lin, Katherine Kairis, Carlisle Turner, and Lori Levin.
\newblock {URIEL} and lang2vec: Representing languages as typological, geographical, and phylogenetic vectors.
\newblock In Mirella Lapata, Phil Blunsom, and Alexander Koller (eds.), \emph{Proceedings of the 15th Conference of the {E}uropean Chapter of the Association for Computational Linguistics: Volume 2, Short Papers}, pp.\  8--14, Valencia, Spain, April 2017. Association for Computational Linguistics.
\newblock URL \url{https://aclanthology.org/E17-2002}.

\bibitem[Lu et~al.(2024)Lu, Zhu, Li, Qiao, and Yuan]{llamax}
Yinquan Lu, Wenhao Zhu, Lei Li, Yu~Qiao, and Fei Yuan.
\newblock Llamax: Scaling linguistic horizons of llm by enhancing translation capabilities beyond 100 languages.
\newblock \emph{arXiv preprint arXiv:2407.05975}, 2024.

\bibitem[Maillard et~al.(2023)Maillard, Gao, Kalbassi, Sadagopan, Goswami, Koehn, Fan, and Guzm{\'a}n]{maillard2023small}
Jean Maillard, Cynthia Gao, Elahe Kalbassi, Kaushik~Ram Sadagopan, Vedanuj Goswami, Philipp Koehn, Angela Fan, and Francisco Guzm{\'a}n.
\newblock Small data, big impact: Leveraging minimal data for effective machine translation.
\newblock In \emph{Proceedings of the 61st Annual Meeting of the Association for Computational Linguistics (Volume 1: Long Papers)}, pp.\  2740--2756, 2023.

\bibitem[Meng et~al.(2024)Meng, Xia, and Chen]{simpo}
Yu~Meng, Mengzhou Xia, and Danqi Chen.
\newblock Simpo: Simple preference optimization with a reference-free reward.
\newblock \emph{arXiv preprint arXiv:2405.14734}, 2024.

\bibitem[OpenAI(2023)]{openai2023gpt4}
OpenAI.
\newblock Gpt-4 technical report, 2023.

\bibitem[{Ortiz Su{'a}rez} et~al.(2019){Ortiz Su{'a}rez}, Sagot, and Romary]{OrtizSuarezSagotRomary2019}
Pedro~Javier {Ortiz Su{'a}rez}, Benoit Sagot, and Laurent Romary.
\newblock Asynchronous pipelines for processing huge corpora on medium to low resource infrastructures.
\newblock Proceedings of the Workshop on Challenges in the Management of Large Corpora (CMLC-7) 2019. Cardiff, 22nd July 2019, pp.\  9 -- 16, Mannheim, 2019. Leibniz-Institut f{"u}r Deutsche Sprache.
\newblock \doi{10.14618/ids-pub-9021}.
\newblock URL \url{http://nbn-resolving.de/urn:nbn:de:bsz:mh39-90215}.

\bibitem[Papineni et~al.(2002)Papineni, Roukos, Ward, and Zhu]{papineni2002bleu}
Kishore Papineni, Salim Roukos, Todd Ward, and Wei-Jing Zhu.
\newblock Bleu: a method for automatic evaluation of machine translation.
\newblock In \emph{Proceedings of the 40th annual meeting of the Association for Computational Linguistics}, pp.\  311--318, 2002.

\bibitem[Petrick et~al.(2023)Petrick, Herold, Petrushkov, Khadivi, and Ney]{petrick-etal-2023-document}
Frithjof Petrick, Christian Herold, Pavel Petrushkov, Shahram Khadivi, and Hermann Ney.
\newblock Document-level language models for machine translation.
\newblock In Philipp Koehn, Barry Haddow, Tom Kocmi, and Christof Monz (eds.), \emph{Proceedings of the Eighth Conference on Machine Translation}, pp.\  375--391, Singapore, December 2023. Association for Computational Linguistics.
\newblock \doi{10.18653/v1/2023.wmt-1.39}.
\newblock URL \url{https://aclanthology.org/2023.wmt-1.39/}.

\bibitem[Post(2018)]{sacrebleu}
Matt Post.
\newblock A call for clarity in reporting {BLEU} scores.
\newblock In \emph{Proceedings of the Third Conference on Machine Translation: Research Papers}, pp.\  186--191, Brussels, Belgium, October 2018. Association for Computational Linguistics.
\newblock \doi{10.18653/v1/W18-6319}.
\newblock URL \url{https://aclanthology.org/W18-6319}.

\bibitem[Rafailov et~al.(2024)Rafailov, Sharma, Mitchell, Manning, Ermon, and Finn]{dpo}
Rafael Rafailov, Archit Sharma, Eric Mitchell, Christopher~D Manning, Stefano Ermon, and Chelsea Finn.
\newblock Direct preference optimization: Your language model is secretly a reward model.
\newblock \emph{Advances in Neural Information Processing Systems}, 36, 2024.

\bibitem[Raffel et~al.(2020)Raffel, Shazeer, Roberts, Lee, Narang, Matena, Zhou, Li, and Liu]{10.5555/3455716.3455856}
Colin Raffel, Noam Shazeer, Adam Roberts, Katherine Lee, Sharan Narang, Michael Matena, Yanqi Zhou, Wei Li, and Peter~J. Liu.
\newblock Exploring the limits of transfer learning with a unified text-to-text transformer.
\newblock \emph{J. Mach. Learn. Res.}, 21\penalty0 (1), January 2020.
\newblock ISSN 1532-4435.

\bibitem[Raunak et~al.(2023)Raunak, Sharaf, Wang, Awadalla, and Menezes]{raunak-etal-2023-leveraging}
Vikas Raunak, Amr Sharaf, Yiren Wang, Hany Awadalla, and Arul Menezes.
\newblock Leveraging {GPT}-4 for automatic translation post-editing.
\newblock In Houda Bouamor, Juan Pino, and Kalika Bali (eds.), \emph{Findings of the Association for Computational Linguistics: EMNLP 2023}, pp.\  12009--12024, Singapore, December 2023. Association for Computational Linguistics.
\newblock \doi{10.18653/v1/2023.findings-emnlp.804}.
\newblock URL \url{https://aclanthology.org/2023.findings-emnlp.804}.

\bibitem[Rei et~al.(2022)Rei, C.~de Souza, Alves, Zerva, Farinha, Glushkova, Lavie, Coheur, and Martins]{comet22}
Ricardo Rei, Jos{\'e}~G. C.~de Souza, Duarte Alves, Chrysoula Zerva, Ana~C Farinha, Taisiya Glushkova, Alon Lavie, Luisa Coheur, and Andr{\'e} F.~T. Martins.
\newblock {COMET}-22: Unbabel-{IST} 2022 submission for the metrics shared task.
\newblock In \emph{Proceedings of the Seventh Conference on Machine Translation (WMT)}, pp.\  578--585, Abu Dhabi, United Arab Emirates (Hybrid), December 2022. Association for Computational Linguistics.
\newblock URL \url{https://aclanthology.org/2022.wmt-1.52}.

\bibitem[Schwenk et~al.(2021)Schwenk, Wenzek, Edunov, Grave, Joulin, and Fan]{schwenk-etal-2021-ccmatrix}
Holger Schwenk, Guillaume Wenzek, Sergey Edunov, Edouard Grave, Armand Joulin, and Angela Fan.
\newblock {CCM}atrix: Mining billions of high-quality parallel sentences on the web.
\newblock In Chengqing Zong, Fei Xia, Wenjie Li, and Roberto Navigli (eds.), \emph{Proceedings of the 59th Annual Meeting of the Association for Computational Linguistics and the 11th International Joint Conference on Natural Language Processing (Volume 1: Long Papers)}, pp.\  6490--6500, Online, August 2021. Association for Computational Linguistics.
\newblock \doi{10.18653/v1/2021.acl-long.507}.
\newblock URL \url{https://aclanthology.org/2021.acl-long.507}.

\bibitem[Shazeer et~al.(2017)Shazeer, Mirhoseini, Maziarz, Davis, Le, Hinton, and Dean]{shazeer2017}
Noam Shazeer, *Azalia Mirhoseini, *Krzysztof Maziarz, Andy Davis, Quoc Le, Geoffrey Hinton, and Jeff Dean.
\newblock Outrageously large neural networks: The sparsely-gated mixture-of-experts layer.
\newblock In \emph{International Conference on Learning Representations}, 2017.
\newblock URL \url{https://openreview.net/forum?id=B1ckMDqlg}.

\bibitem[Singh et~al.(2024)Singh, Vargus, Dsouza, Karlsson, Mahendiran, Ko, Shandilya, Patel, Mataciunas, OMahony, et~al.]{aya_data}
Shivalika Singh, Freddie Vargus, Daniel Dsouza, B{\"o}rje~F Karlsson, Abinaya Mahendiran, Wei-Yin Ko, Herumb Shandilya, Jay Patel, Deividas Mataciunas, Laura OMahony, et~al.
\newblock Aya dataset: An open-access collection for multilingual instruction tuning.
\newblock \emph{arXiv preprint arXiv:2402.06619}, 2024.

\bibitem[Team et~al.(2024{\natexlab{a}})Team, Mesnard, Hardin, Dadashi, Bhupatiraju, Pathak, Sifre, Rivi{\`e}re, Kale, Love, et~al.]{gemma1}
Gemma Team, Thomas Mesnard, Cassidy Hardin, Robert Dadashi, Surya Bhupatiraju, Shreya Pathak, Laurent Sifre, Morgane Rivi{\`e}re, Mihir~Sanjay Kale, Juliette Love, et~al.
\newblock Gemma: Open models based on gemini research and technology.
\newblock \emph{arXiv preprint arXiv:2403.08295}, 2024{\natexlab{a}}.

\bibitem[Team et~al.(2024{\natexlab{b}})Team, Riviere, Pathak, Sessa, Hardin, Bhupatiraju, Hussenot, Mesnard, Shahriari, Ram{\'e}, et~al.]{gemma2}
Gemma Team, Morgane Riviere, Shreya Pathak, Pier~Giuseppe Sessa, Cassidy Hardin, Surya Bhupatiraju, L{\'e}onard Hussenot, Thomas Mesnard, Bobak Shahriari, Alexandre Ram{\'e}, et~al.
\newblock Gemma 2: Improving open language models at a practical size.
\newblock \emph{arXiv preprint arXiv:2408.00118}, 2024{\natexlab{b}}.

\bibitem[Team et~al.(2022)Team, Costa-jussà, Cross, Çelebi, Elbayad, Heafield, Heffernan, Kalbassi, Lam, Licht, Maillard, Sun, Wang, Wenzek, Youngblood, Akula, Barrault, Gonzalez, Hansanti, Hoffman, Jarrett, Sadagopan, Rowe, Spruit, Tran, Andrews, Ayan, Bhosale, Edunov, Fan, Gao, Goswami, Guzmán, Koehn, Mourachko, Ropers, Saleem, Schwenk, and Wang]{nllb}
NLLB Team, Marta~R. Costa-jussà, James Cross, Onur Çelebi, Maha Elbayad, Kenneth Heafield, Kevin Heffernan, Elahe Kalbassi, Janice Lam, Daniel Licht, Jean Maillard, Anna Sun, Skyler Wang, Guillaume Wenzek, Al~Youngblood, Bapi Akula, Loic Barrault, Gabriel~Mejia Gonzalez, Prangthip Hansanti, John Hoffman, Semarley Jarrett, Kaushik~Ram Sadagopan, Dirk Rowe, Shannon Spruit, Chau Tran, Pierre Andrews, Necip~Fazil Ayan, Shruti Bhosale, Sergey Edunov, Angela Fan, Cynthia Gao, Vedanuj Goswami, Francisco Guzmán, Philipp Koehn, Alexandre Mourachko, Christophe Ropers, Safiyyah Saleem, Holger Schwenk, and Jeff Wang.
\newblock No language left behind: Scaling human-centered machine translation.
\newblock 2022.
\newblock URL \url{https://arxiv.org/abs/2207.04672}.

\bibitem[Thompson \& Post(2020{\natexlab{a}})Thompson and Post]{thompson-post-2020-automatic}
Brian Thompson and Matt Post.
\newblock Automatic machine translation evaluation in many languages via zero-shot paraphrasing.
\newblock In \emph{Proceedings of the 2020 Conference on Empirical Methods in Natural Language Processing (EMNLP)}, pp.\  90--121, Online, November 2020{\natexlab{a}}. Association for Computational Linguistics.
\newblock \doi{10.18653/v1/2020.emnlp-main.8}.
\newblock URL \url{https://aclanthology.org/2020.emnlp-main.8/}.

\bibitem[Thompson \& Post(2020{\natexlab{b}})Thompson and Post]{thompson-post-2020-paraphrase}
Brian Thompson and Matt Post.
\newblock Paraphrase generation as zero-shot multilingual translation: Disentangling semantic similarity from lexical and syntactic diversity.
\newblock In \emph{Proceedings of the Fifth Conference on Machine Translation}, pp.\  561--570, Online, November 2020{\natexlab{b}}. Association for Computational Linguistics.
\newblock URL \url{https://aclanthology.org/2020.wmt-1.67/}.

\bibitem[Thompson et~al.(2024)Thompson, Dhaliwal, Frisch, Domhan, and Federico]{thompson-etal-2024-shocking}
Brian Thompson, Mehak Dhaliwal, Peter Frisch, Tobias Domhan, and Marcello Federico.
\newblock A shocking amount of the web is machine translated: Insights from multi-way parallelism.
\newblock In Lun-Wei Ku, Andre Martins, and Vivek Srikumar (eds.), \emph{Findings of the Association for Computational Linguistics: ACL 2024}, pp.\  1763--1775, Bangkok, Thailand, August 2024. Association for Computational Linguistics.
\newblock \doi{10.18653/v1/2024.findings-acl.103}.
\newblock URL \url{https://aclanthology.org/2024.findings-acl.103/}.

\bibitem[Tiedemann(2012)]{tiedemann-2012-parallel}
J{\"o}rg Tiedemann.
\newblock Parallel data, tools and interfaces in {OPUS}.
\newblock In Nicoletta Calzolari, Khalid Choukri, Thierry Declerck, Mehmet~U{\u{g}}ur Do{\u{g}}an, Bente Maegaard, Joseph Mariani, Asuncion Moreno, Jan Odijk, and Stelios Piperidis (eds.), \emph{Proceedings of the Eighth International Conference on Language Resources and Evaluation ({LREC}'12)}, pp.\  2214--2218, Istanbul, Turkey, May 2012. European Language Resources Association (ELRA).
\newblock URL \url{http://www.lrec-conf.org/proceedings/lrec2012/pdf/463_Paper.pdf}.

\bibitem[Touvron et~al.(2023{\natexlab{a}})Touvron, Lavril, Izacard, Martinet, Lachaux, Lacroix, Rozi{\`e}re, Goyal, Hambro, Azhar, et~al.]{llama1}
Hugo Touvron, Thibaut Lavril, Gautier Izacard, Xavier Martinet, Marie-Anne Lachaux, Timoth{\'e}e Lacroix, Baptiste Rozi{\`e}re, Naman Goyal, Eric Hambro, Faisal Azhar, et~al.
\newblock Llama: Open and efficient foundation language models.
\newblock \emph{arXiv preprint arXiv:2302.13971}, 2023{\natexlab{a}}.

\bibitem[Touvron et~al.(2023{\natexlab{b}})Touvron, Martin, Stone, Albert, Almahairi, Babaei, Bashlykov, Batra, Bhargava, Bhosale, et~al.]{llama2}
Hugo Touvron, Louis Martin, Kevin Stone, Peter Albert, Amjad Almahairi, Yasmine Babaei, Nikolay Bashlykov, Soumya Batra, Prajjwal Bhargava, Shruti Bhosale, et~al.
\newblock Llama 2: Open foundation and fine-tuned chat models.
\newblock \emph{arXiv preprint arXiv:2307.09288}, 2023{\natexlab{b}}.

\bibitem[{\"U}st{\"u}n et~al.(2024){\"U}st{\"u}n, Aryabumi, Yong, Ko, D'souza, Onilude, Bhandari, Singh, Ooi, Kayid, et~al.]{aya101}
Ahmet {\"U}st{\"u}n, Viraat Aryabumi, Zheng-Xin Yong, Wei-Yin Ko, Daniel D'souza, Gbemileke Onilude, Neel Bhandari, Shivalika Singh, Hui-Lee Ooi, Amr Kayid, et~al.
\newblock Aya model: An instruction finetuned open-access multilingual language model.
\newblock \emph{arXiv preprint arXiv:2402.07827}, 2024.

\bibitem[Wang et~al.(2021)Wang, Tsvetkov, Firat, and Cao]{wang2021gradient}
Zirui Wang, Yulia Tsvetkov, Orhan Firat, and Yuan Cao.
\newblock Gradient vaccine: Investigating and improving multi-task optimization in massively multilingual models.
\newblock In \emph{International Conference on Learning Representations}, 2021.
\newblock URL \url{https://openreview.net/forum?id=F1vEjWK-lH_}.

\bibitem[Xu et~al.(2023)Xu, Tan, Li, Chen, Van~Durme, Koehn, and Murray]{xu-etal-2023-condensing}
Haoran Xu, Weiting Tan, Shuyue Li, Yunmo Chen, Benjamin Van~Durme, Philipp Koehn, and Kenton Murray.
\newblock Condensing multilingual knowledge with lightweight language-specific modules.
\newblock In Houda Bouamor, Juan Pino, and Kalika Bali (eds.), \emph{Proceedings of the 2023 Conference on Empirical Methods in Natural Language Processing}, pp.\  1575--1587, Singapore, December 2023. Association for Computational Linguistics.
\newblock \doi{10.18653/v1/2023.emnlp-main.97}.
\newblock URL \url{https://aclanthology.org/2023.emnlp-main.97}.

\bibitem[Xu et~al.(2024{\natexlab{a}})Xu, Kim, Sharaf, and Awadalla]{alma}
Haoran Xu, Young~Jin Kim, Amr Sharaf, and Hany~Hassan Awadalla.
\newblock A paradigm shift in machine translation: Boosting translation performance of large language models.
\newblock In \emph{The Twelfth International Conference on Learning Representations}, 2024{\natexlab{a}}.
\newblock URL \url{https://openreview.net/forum?id=farT6XXntP}.

\bibitem[Xu et~al.(2024{\natexlab{b}})Xu, Sharaf, Chen, Tan, Shen, Van~Durme, Murray, and Kim]{alma-r}
Haoran Xu, Amr Sharaf, Yunmo Chen, Weiting Tan, Lingfeng Shen, Benjamin Van~Durme, Kenton Murray, and Young~Jin Kim.
\newblock Contrastive preference optimization: Pushing the boundaries of llm performance in machine translation.
\newblock In \emph{Forty-first International Conference on Machine Learning}, 2024{\natexlab{b}}.

\bibitem[Xue et~al.(2020)Xue, Constant, Roberts, Kale, Al-Rfou, Siddhant, Barua, and Raffel]{mt5}
Linting Xue, Noah Constant, Adam Roberts, Mihir Kale, Rami Al-Rfou, Aditya Siddhant, Aditya Barua, and Colin Raffel.
\newblock mt5: A massively multilingual pre-trained text-to-text transformer.
\newblock \emph{arXiv preprint arXiv:2010.11934}, 2020.

\bibitem[Xue et~al.(2021)Xue, Constant, Roberts, Kale, Al-Rfou, Siddhant, Barua, and Raffel]{xue-etal-2021-mt5}
Linting Xue, Noah Constant, Adam Roberts, Mihir Kale, Rami Al-Rfou, Aditya Siddhant, Aditya Barua, and Colin Raffel.
\newblock m{T}5: A massively multilingual pre-trained text-to-text transformer.
\newblock In Kristina Toutanova, Anna Rumshisky, Luke Zettlemoyer, Dilek Hakkani-Tur, Iz~Beltagy, Steven Bethard, Ryan Cotterell, Tanmoy Chakraborty, and Yichao Zhou (eds.), \emph{Proceedings of the 2021 Conference of the North American Chapter of the Association for Computational Linguistics: Human Language Technologies}, pp.\  483--498, Online, June 2021. Association for Computational Linguistics.
\newblock \doi{10.18653/v1/2021.naacl-main.41}.
\newblock URL \url{https://aclanthology.org/2021.naacl-main.41/}.

\bibitem[Yang et~al.(2023)Yang, Li, Zhang, and Zong]{bigtranslate}
Wen Yang, Chong Li, Jiajun Zhang, and Chengqing Zong.
\newblock Bigtrans: Augmenting large language models with multilingual translation capability over 100 languages.
\newblock \emph{arXiv preprint arXiv:2305.18098}, 2023.

\bibitem[Zeng et~al.(2023)Zeng, Meng, Yin, and Zhou]{zeng2023tim}
Jiali Zeng, Fandong Meng, Yongjing Yin, and Jie Zhou.
\newblock Tim: Teaching large language models to translate with comparison.
\newblock \emph{arXiv preprint arXiv:2307.04408}, 2023.

\bibitem[Zhang et~al.(2020)Zhang, Williams, Titov, and Sennrich]{improvingmmt}
Biao Zhang, Philip Williams, Ivan Titov, and Rico Sennrich.
\newblock Improving massively multilingual neural machine translation and zero-shot translation.
\newblock In \emph{Proceedings of the 58th Annual Meeting of the Association for Computational Linguistics}, pp.\  1628--1639, Online, July 2020. Association for Computational Linguistics.
\newblock \doi{10.18653/v1/2020.acl-main.148}.
\newblock URL \url{https://aclanthology.org/2020.acl-main.148}.

\bibitem[Zhang et~al.(2023)Zhang, Fang, Zhang, Ma, Zhou, Huang, Bu, Gui, Chen, Chen, et~al.]{bayling}
Shaolei Zhang, Qingkai Fang, Zhuocheng Zhang, Zhengrui Ma, Yan Zhou, Langlin Huang, Mengyu Bu, Shangtong Gui, Yunji Chen, Xilin Chen, et~al.
\newblock Bayling: Bridging cross-lingual alignment and instruction following through interactive translation for large language models.
\newblock \emph{arXiv preprint arXiv:2306.10968}, 2023.

\bibitem[Zhang et~al.(2022)Zhang, Roller, Goyal, Artetxe, Chen, Chen, Dewan, Diab, Li, Lin, et~al.]{OPT}
Susan Zhang, Stephen Roller, Naman Goyal, Mikel Artetxe, Moya Chen, Shuohui Chen, Christopher Dewan, Mona Diab, Xian Li, Xi~Victoria Lin, et~al.
\newblock Opt: Open pre-trained transformer language models.
\newblock \emph{arXiv preprint arXiv:2205.01068}, 2022.

\bibitem[Zhu et~al.(2024{\natexlab{a}})Zhu, Trenous, Shen, Klakow, Byrne, and Hasler]{zhu-etal-2024-preference}
Dawei Zhu, Sony Trenous, Xiaoyu Shen, Dietrich Klakow, Bill Byrne, and Eva Hasler.
\newblock A preference-driven paradigm for enhanced translation with large language models.
\newblock In Kevin Duh, Helena Gomez, and Steven Bethard (eds.), \emph{Proceedings of the 2024 Conference of the North American Chapter of the Association for Computational Linguistics: Human Language Technologies (Volume 1: Long Papers)}, pp.\  3385--3403, Mexico City, Mexico, June 2024{\natexlab{a}}. Association for Computational Linguistics.
\newblock \doi{10.18653/v1/2024.naacl-long.186}.
\newblock URL \url{https://aclanthology.org/2024.naacl-long.186}.

\bibitem[Zhu et~al.(2024{\natexlab{b}})Zhu, Liu, Dong, Xu, Huang, Kong, Chen, and Li]{zhu-etal-2024-multilingual}
Wenhao Zhu, Hongyi Liu, Qingxiu Dong, Jingjing Xu, Shujian Huang, Lingpeng Kong, Jiajun Chen, and Lei Li.
\newblock Multilingual machine translation with large language models: Empirical results and analysis.
\newblock In Kevin Duh, Helena Gomez, and Steven Bethard (eds.), \emph{Findings of the Association for Computational Linguistics: NAACL 2024}, pp.\  2765--2781, Mexico City, Mexico, June 2024{\natexlab{b}}. Association for Computational Linguistics.
\newblock \doi{10.18653/v1/2024.findings-naacl.176}.
\newblock URL \url{https://aclanthology.org/2024.findings-naacl.176}.

\bibitem[Zouhar et~al.(2024)Zouhar, Ding, Currey, Badeka, Wang, and Thompson]{zouhar-etal-2024-fine}
Vil{\'e}m Zouhar, Shuoyang Ding, Anna Currey, Tatyana Badeka, Jenyuan Wang, and Brian Thompson.
\newblock Fine-tuned machine translation metrics struggle in unseen domains.
\newblock In Lun-Wei Ku, Andre Martins, and Vivek Srikumar (eds.), \emph{Proceedings of the 62nd Annual Meeting of the Association for Computational Linguistics (Volume 2: Short Papers)}, pp.\  488--500, Bangkok, Thailand, August 2024. Association for Computational Linguistics.
\newblock \doi{10.18653/v1/2024.acl-short.45}.
\newblock URL \url{https://aclanthology.org/2024.acl-short.45/}.

\end{thebibliography}
\bibliographystyle{iclr2025_conference}

\newpage
\appendix
\section*{Appendix Contents}
\appendix
\begin{table}[ht]
    \centering
    \footnotesize
    \begin{tabular}{cl}
    \textbf{Appendix Sections}    & \textbf{Contents}  \\ \toprule
    \autoref{app:sec:lg_info} &  \begin{tabular}[c]{@{}l@{}} Language Information \end{tabular} \\ \midrule
    \autoref{app:sec:training_recipe} &  \begin{tabular}[c]{@{}l@{}} Illustration of Training Recipe \end{tabular} \\ \midrule
     \autoref{app:cdf} &  \begin{tabular}[c]{@{}l@{}} Reward Differences of MT and QA \end{tabular} \\ \midrule
    \autoref{app:sec:prompt}     &  \begin{tabular}[c]{@{}l@{}} Prompts \end{tabular} \\ \midrule
     \autoref{app:sec:full_results}     &  \begin{tabular}[c]{@{}l@{}} Full Results of All Directions\end{tabular} \\ \midrule
     \autoref{app:sec:over_reject}     &  \begin{tabular}[c]{@{}l@{}} Examples of Over-Rejection \end{tabular} \\
    \bottomrule
\end{tabular}    
\end{table}

\section{Language Information}
\label{app:sec:lg_info}
We provide detailed information on the eight language groups, including their scripts, language families, and resource levels, in Table \ref{app:tab:lang_info}. Each group includes English to ensure that each language-specific module supports English-centric translation and to prevent catastrophic forgetting of English. While we primarily grouped languages based on linguistic similarity, the grouping is not perfect. This is due to the need to balance the number of languages in each group and the inherent nature of language resources. For example, Group 6 is a mix of Asian and European languages, and although most languages in Group 4 are Southeast Asian languages, we include French as an additional bonus language to facilitate cross-lingual transfer, especially since most languages in this group are low- and mid-resource. 

\begin{longtable}{cccccc}
\caption{Detailed information of all langauges.} 
\label{app:tab:lang_info} \\
\hline
Language    & ISO-639-1 & Script       & Family        & Subgroup          & Resource \\ \hline
\endfirsthead
\hline
Language    & ISO-639-1 & Script       & Family        & Subgroup          & Resource \\ \hline
\endhead
\hline
\multicolumn{6}{r}{\textit{Continued on next page}} \\
\endfoot
\hline
\endlastfoot
English     & en        & Latin        & Indo-European & Germanic          & High     \\ \hline
\multicolumn{6}{l}{\textit{Group 1: Germanic languages}}                              \\ \hline
Afrikaans   & af        & Latin        & Indo-European & Germanic          & Mid      \\
Danish      & da        & Latin        & Indo-European & Germanic          & Mid      \\
Dutch       & nl        & Latin        & Indo-European & Germanic          & High     \\
German      & de        & Latin        & Indo-European & Germanic          & High     \\
Icelandic   & is        & Latin        & Indo-European & Germanic          & Low      \\
Norwegian   & no        & Latin        & Indo-European & Germanic          & Low      \\
Swedish     & sv        & Latin        & Indo-European & Germanic          & High     \\ \hline
\multicolumn{6}{l}{\textit{Group 2: Romance Languages}}                               \\ \hline
Catalan     & ca        & Latin        & Indo-European & Italic            & High     \\
Galician    & gl        & Latin        & Indo-European & Italic            & Mid      \\
Italian     & it        & Latin        & Indo-European & Italic            & High     \\
Portuguese  & pt        & Latin        & Indo-European & Italic            & High     \\
Romanian    & ro        & Latin        & Indo-European & Italic            & Mid      \\
Spanish     & es        & Latin        & Indo-European & Italic            & High     \\ \hline
\multicolumn{6}{l}{\textit{Group 3: Eastern and Southern Slavic Languages}}           \\ \hline
Bulgarian   & bg        & Cyrillic     & Indo-European & Balto-Slavic      & Mid      \\
Macedonian  & mk        & Cyrillic     & Indo-European & Balto-Slavic      & Low      \\
Russian     & ru        & Cyrillic     & Indo-European & Balto-Slavic      & High     \\
Serbian     & sr        & Cyrillic     & Indo-European & Balto-Slavic      & High     \\
Ukrainian   & uk        & Cyrillic     & Indo-European & Balto-Slavic      & Mid      \\ \hline
\multicolumn{6}{l}{\textit{Group 4: Southeast Asian Languages}}                       \\ \hline
French      & fr        & Latin        & Indo-European & Italic            & High     \\
Indonesian  & id        & Latin        & Austronesian  & Malayo-Polynesian & Mid      \\
Malagasy    & mg        & Latin        & Austronesian  & Malayo-Polynesian & Low      \\
Malay       & ms        & Latin        & Austronesian  & Malayo-Polynesian & Mid      \\
Thai        & th        & Thai         & Tai-Kadai     & Kam-Tai           & Mid      \\
Vietnamese  & vi        & Latin        & Austronesian  & Vietic            & High     \\ \hline
\multicolumn{6}{l}{\textit{Group 5: Central and Eastern European Languages}}          \\ \hline
Czech       & cs        & Latin        & Indo-European & Balto-Slavic      & High     \\
Greek       & el        & Greek        & Indo-European & Graeco-Phrygian   & Mid      \\
Hungarian   & hu        & Latin        & Uralic        & Finnic            & High     \\
Latvian     & lv        & Latin        & Indo-European & Balto-Slavic      & Mid      \\
Lithuanian  & lt        & Latin        & Indo-European & Balto-Slavic      & Mid      \\
Polish      & pl        & Latin        & Indo-European & Balto-Slavic      & High     \\ \hline
\multicolumn{6}{l}{\textit{Group 6: Eurasian Language Mix}}                           \\ \hline
Chinese     & zh        & Han          & Sino-Tibetan  & Sinitic           & High     \\
Estonian    & et        & Latin        & Uralic        & Finnic            & Mid      \\
Finnish     & fi        & Latin        & Uralic        & Finnic            & High     \\
Georgian    & ka        & Georgian     & Kartvelian    & Georgian-Zan      & Mid      \\
Japanese    & ja        & Japanese     & Japonic       & Japanesic         & High     \\
Korean      & ko        & Hangul       & Koreanic      & Korean            & High     \\ \hline
\multicolumn{6}{l}{\textit{Group 7: Indo-Aryan Languages}}                            \\ \hline
Gujarati    & gu        & Gujarati     & Indo-European & Indo-Aryan        & Low      \\
Hindi       & hi        & Devanagari   & Indo-European & Indo-Aryan        & High     \\
Marathi     & mr        & Devanagari   & Indo-European & Indo-Aryan        & Low      \\
Nepali      & ne        & Devanagari   & Indo-European & Indo-Aryan        & Low      \\
Urdu        & ur        & Arabic       & Indo-European & Indo-Aryan        & Mid      \\ \hline
\multicolumn{6}{l}{\textit{Group 8: Turkic and Semitic Languages}}                    \\ \hline
Arabic      & ar        & Arabic       & Afro-Asiatic  & Semitic           & High     \\
Azerbaijani & az        & Arabic/Latin & Turkic        & Common Turkic     & Low      \\
Hebrew      & he        & Hebrew       & Afro-Asiatic  & Semitic           & Mid      \\
Kazakh      & kk        & Cyrillic     & Turkic        & Common Turkic     & Mid      \\
Kyrgyz      & ky        & Cyrillic     & Turkic        & Common Turkic     & Low      \\
Persian     & fa        & Arabic       & Indo-European & Iranian           & High     \\
Turkish     & tr        & Latin        & Turkic        & Common Turkic     & High     \\
Uzbek       & uz        & Latin        & Turkic        & Common Turkic     & Low      \\ \hline
\end{longtable}

\section{Illustration of Training Recipe}
\label{app:sec:training_recipe}
\begin{figure*}[ht]
    \centering
    \resizebox{1\linewidth}{!}{
    \includegraphics[width=7.5cm]{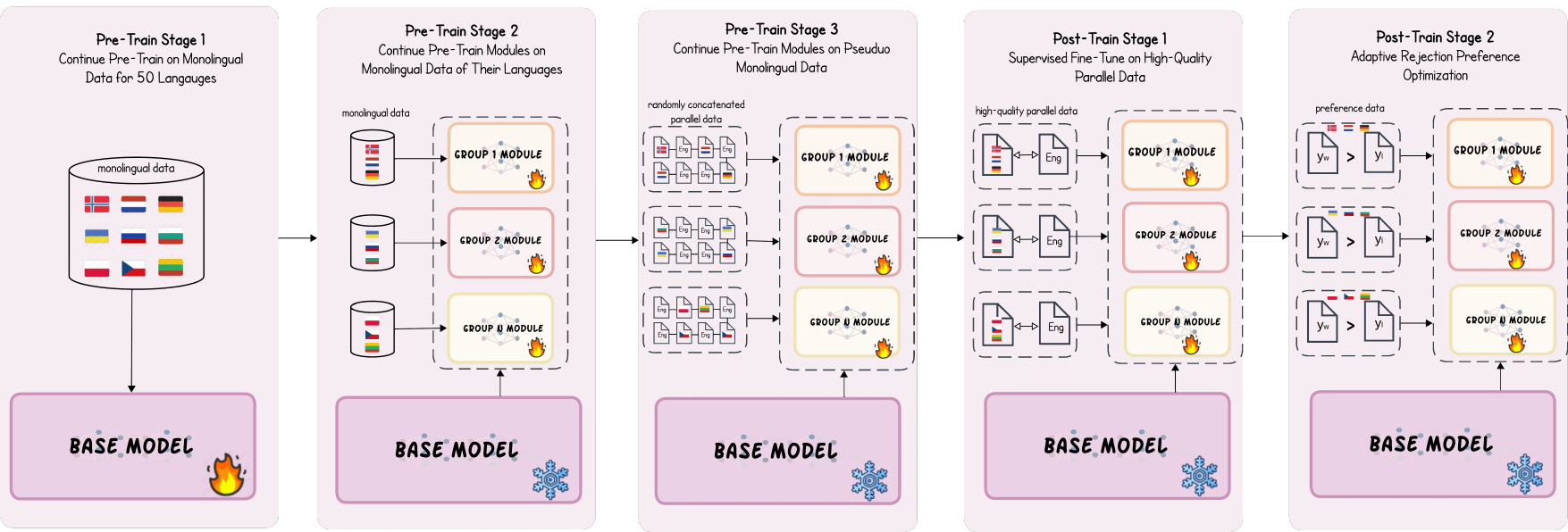}}
    \caption{
    This diagram of the multi-stage process of fine-tuning a multilingual model. \textit{In Pre-Training Stage 1}, the base model is fine-tuned using 20B tokens of monolingual data from 50 languages. The process continues with \textit{Pre-Training Stage 2}, where language-specific modules are fine-tuned with 10B monolingual tokens. \textit{Pre-Training Stage 3} introduces pseudo-monolingual fine-tuning, using randomly concatenated parallel sentences to improve multilingual alignment. The model then undergoes \textit{Post-Training Stage 1}, where SFT is performed on high-quality parallel data, followed by \textit{Post-Training Stage 2}, which applies Adaptive Contrastive Preference Optimization to address over-rejection issues in translation preference learning.
    }
    \label{app:fig:pipeline}
\end{figure*}

Here, we illustrate an overview of our 5-step training recipe in Figure \ref{app:fig:pipeline}. 

\section{Reward Differences of MT and QA}
\label{app:cdf}
\begin{figure}[h]
    \centering
    \resizebox{0.6\linewidth}{!}{
    \includegraphics[width=7.5cm]{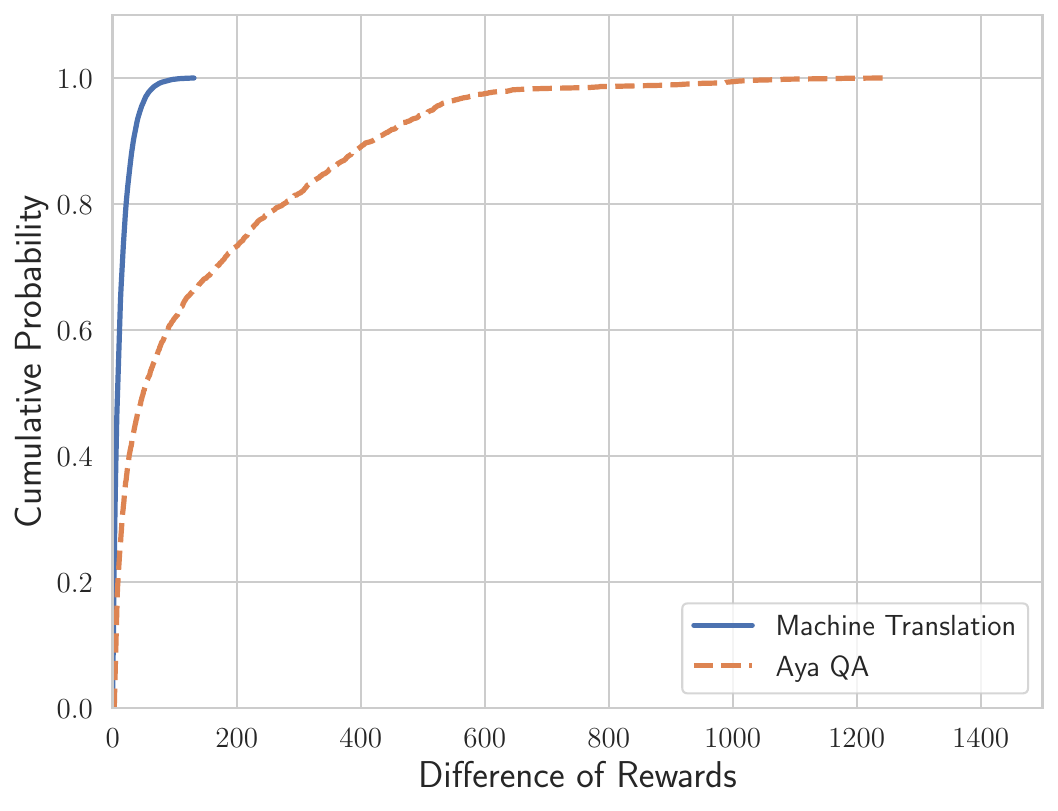}}
    \caption{
        Cumulative distribution of reward differences between machine translation and open-ended question answering tasks in contrastive preference optimization. 
        %The reward differences are sorted and presented as cumulative probabilities. The open-ended QA task demonstrates significantly larger reward differences compared to the MT task, as reflected in the wider spread of the Aya QA curve.
    }
    \label{app:fig:cdf}
\end{figure}

Here, we present a comparison of the reward difference between machine translation (MT) tasks and open-ended question answering (QA) tasks in preference learning. Figure \ref{app:fig:cdf} illustrates the cumulative distribution of reward differences for the MT preference dataset, as described in Section \ref{sec:experiment}, alongside the multilingual preference data from the Aya open-ended QA dataset \citep{aya_data} for languages in Group 6. The reward differences are sorted in ascending order, and their cumulative probabilities are displayed. The reward difference is computed using the CPO loss function: $\log \pi_{\theta}(y_w | x) - \log \pi_{\theta}(y_l | x)$. The construction of the Aya preference dataset follows the same methodology as the MT preference data, where we fine-tune the Aya QA dataset via SFT and use the fine-tuned model to generate answers for the training data. System-generated responses are treated as dis-preferred, while original references are considered preferred. As shown in Figure \ref{app:fig:cdf}, the open-ended QA task exhibits significantly larger reward differences compared to machine translation. For instance, the maximum reward difference for the smallest 80\% of MT preference data is 20, whereas it is approximately 300 for Aya QA. Similarly, the maximum reward difference for the MT preference data is 131, while that for Aya QA is nearly tenfold larger.

\section{Prompts}
\label{app:sec:prompt}
\begin{figure}[h]
    \centering
    \resizebox{0.85\linewidth}{!}{
    \includegraphics[width=7.5cm]{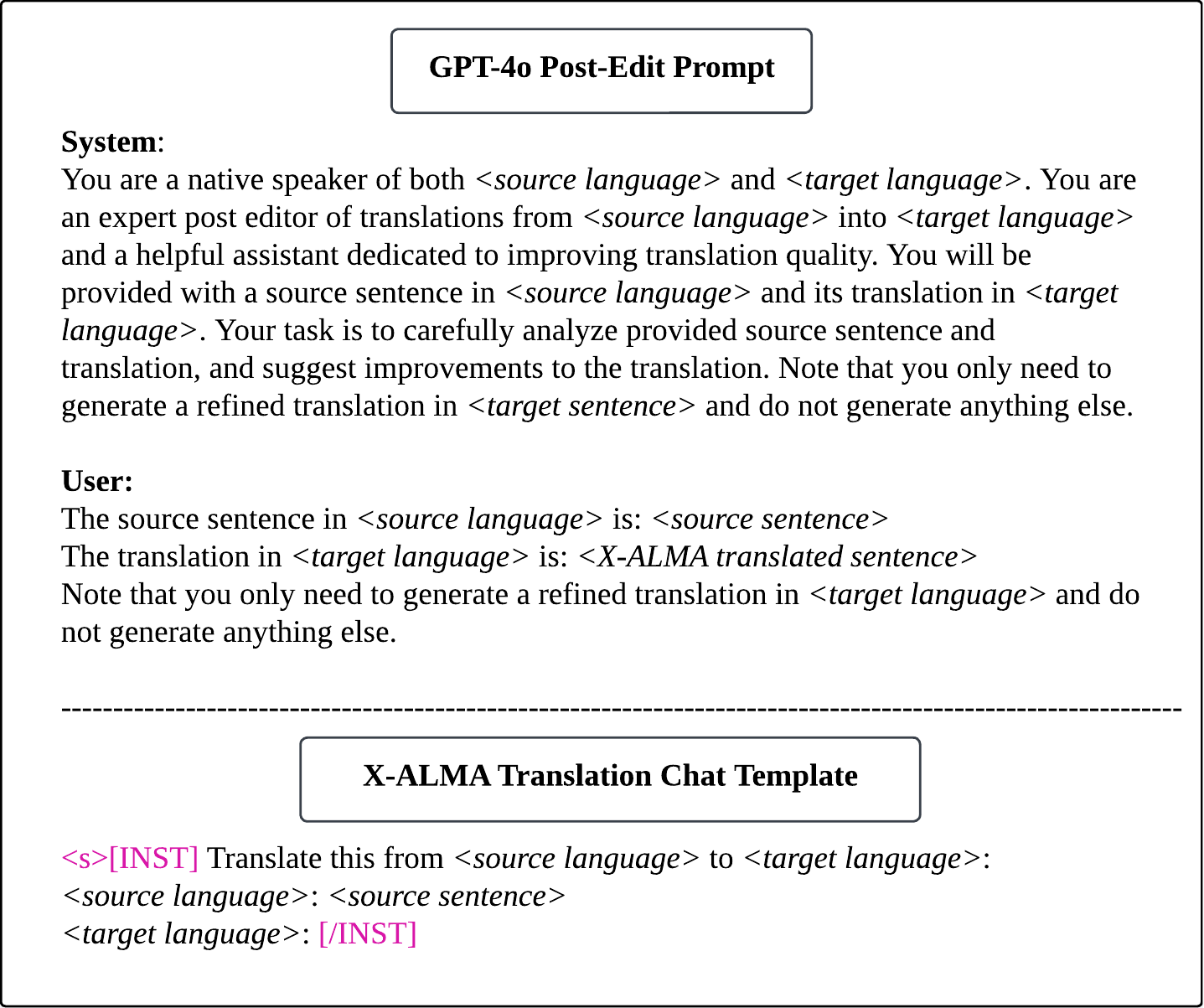}}
    \caption{
    Prompts used for GPT-4o post editing and X-ALMA translation generation.
    }
    \label{app:fig:prompts}
\end{figure}

In Figure \ref{app:fig:prompts}, we present the prompt used for GPT-4o post-editing during the construction of the preference dataset, as well as the prompt used for X-ALMA in generating translations.

\section{Full results}
\label{app:sec:full_results}

We report translation quality using XCOMET-XL, as recent WMT metric shared tasks \citep{freitag-etal-2023-results, freitag-etal-2024-llms} have found high correlation between trained metrics like XCOMET-XL and human preferences. 
However, those findings are limited to a few  languages, and correlation with human judgments has also been shown to degrade for trained metrics in out of domain (relative to WMT, i.e. FLORES) settings \citep{zouhar-etal-2024-fine}. For these reasons we also report the more traditional BLEU metric.

Tables \ref{app:tab:full_group1} to \ref{app:tab:full_group8} present the results for each translation direction across language groups in the Flores-200 dataset, while Table \ref{app:tab:full_wmt23} shows the full results for the WMT'23 dataset. On the Flores-200 dataset, X-ALMA surpasses all other open-source multilingual models in every translation direction according to COMET-22, and in 97 out of 98 directions according to XCOMET-XL. Additionally, ARPO, when compared to SFT, demonstrates superior performance in all translation directions reported by COMET-22 and in 95 out of 98 directions according to XCOMET-XL.

\begin{table}[h]
\caption{
Full results for Group 1 in the Flores test data.
}
\vskip 0.05in
\label{app:tab:full_group1}
\centering

\resizebox{1\linewidth}{!}{
\begin{tabular}{lcccccccccccc}
\hline
\multicolumn{1}{c}{\multirow{2}{*}{Models}} & \multicolumn{3}{c}{\texttt{en}$\rightarrow$\texttt{af}} & \multicolumn{3}{c}{\texttt{en}$\rightarrow$\texttt{da}} & \multicolumn{3}{c}{\texttt{en}$\rightarrow$\texttt{de}} & \multicolumn{3}{c}{\texttt{en}$\rightarrow$\texttt{is}}      \\
\cmidrule(lr){2-4} \cmidrule(lr){5-7} \cmidrule(lr){8-10}  \cmidrule(lr){11-13} 
\multicolumn{1}{c}{}                        & BLEU                         & COMET-22                     & XCOMET                   & BLEU                         & COMET-22                     & XCOMET                   & BLEU                   & COMET-22               & \multicolumn{1}{c}{XCOMET}          & BLEU                           & COMET-22                       & XCOMET                    \\ \hline
LLaMA-3.1-8B-Instruct                                  & 34.4                         & 84.8                         & 74.5                        & 37.3                         & 86.6                         & 71.7                        & 30.2                   & 77.5                   & 68.8                                    & 11.9                           & 69.4                           & 53.6                         \\
NLLB-3.3B                                   & 38.9                         & 87.4                         & 74.8                        & 44.5                         & 90.0                         & 76.8                        & 40.0                   & 88.1                   & 76.2                                    & 24.5                           & 84.6                           & 74.1                         \\
\hline
\end{tabular}
}
\end{table}

\begin{table}[h]
\caption{
Full results for Group 2 in the Flores test data.
}
\vskip 0.05in
\label{app:tab:full_group2}
\centering

\resizebox{1\linewidth}{!}{
\begin{tabular}{lcccccccccccc}
\hline
\multicolumn{1}{c}{\multirow{2}{*}{Models}} & \multicolumn{3}{c}{\texttt{en}$\rightarrow$\texttt{ca}} & \multicolumn{3}{c}{\texttt{en}$\rightarrow$\texttt{es}} & \multicolumn{3}{c}{\texttt{en}$\rightarrow$\texttt{gl}}               & \multicolumn{3}{c}{\texttt{en}$\rightarrow$\texttt{it}} \\
\cmidrule(lr){2-4} \cmidrule(lr){5-7} \cmidrule(lr){8-10} \cmidrule(lr){11-13}
\multicolumn{1}{c}{}                        & BLEU                         & COMET-22                     & XCOMET                      & BLEU                         & COMET-22                     & XCOMET                      & BLEU                              & COMET-22                          & XCOMET                          & BLEU                         & COMET-22                     & XCOMET                      \\ \hline
Llama-3.1                                   & 37.5                         & 86.8                         & 77.8                        & 25.0                         & 83.9                         & 77.2                        & 30.2                              & 84.3                              & 74.0                            & 24.9                         & 81.7                         & 72.8                        \\
NLLB-3.3B                                   & 43.1                         & 87.8                         & 77.6                        & 28.6                         & 86.5                         & 80.0                        & 35.7                              & 87.3                              & 76.8                            & \textbf{31.3}                         & 88.5                         & 80.5                        \\
\hline
\end{tabular}
}
\end{table}

\begin{table}[h]
\caption{
Full results for Group 3 in the Flores test data.
}
\vskip 0.05in
\label{app:tab:full_group3}
\centering

\resizebox{1\linewidth}{!}{
\begin{tabular}{lcccccccccccc}
\hline
\multicolumn{1}{c}{\multirow{2}{*}{Models}} & \multicolumn{3}{c}{\texttt{en}$\rightarrow$\texttt{bg}} & \multicolumn{3}{c}{\texttt{en}$\rightarrow$\texttt{mk}}      & \multicolumn{3}{c}{\texttt{en}$\rightarrow$\texttt{ru}} & \multicolumn{3}{c}{\texttt{en}$\rightarrow$\texttt{sr}} \\
\cmidrule(lr){2-4} \cmidrule(lr){5-7} \cmidrule(lr){8-10} \cmidrule(lr){11-13}
\multicolumn{1}{c}{}                        & BLEU                         & COMET-22                     & XCOMET                      & BLEU                           & COMET-22                       & XCOMET                       & BLEU                         & COMET-22                     & XCOMET                      & BLEU                         & COMET-22                     & XCOMET                      \\ \hline
Llama-3.1                                   & 29.6                         & 87.8                         & 74.2                        & 24.9                           & 85.7                           & 74.3                         & 14.4                         & 63.0                         & 45.1                        & 1.4                          & 75.3                         & 75.4                        \\
NLLB-3.3B                                   & 40.5                         & 90.9                         & 77.5                        & 34.4                           & 88.8                           & 77.4                         & 32.2                         & 89.2                         & 77.5                        & 33.8                         & 89.0                         & 77.1                        \\
\hline
\end{tabular}
}
\end{table}

\begin{table}[h]
\caption{
Full results for Group 4 in the Flores test data.
}
\vskip 0.05in
\label{app:tab:full_group4}
\centering

\resizebox{1\linewidth}{!}{
\begin{tabular}{lcccccccccccc}
\hline
\multicolumn{1}{c}{\multirow{2}{*}{Models}} & \multicolumn{3}{c}{\texttt{en}$\rightarrow$\texttt{fr}} & \multicolumn{3}{c}{\texttt{en}$\rightarrow$\texttt{id}} & \multicolumn{3}{c}{\texttt{en}$\rightarrow$\texttt{mg}}      & \multicolumn{3}{c}{\texttt{en}$\rightarrow$\texttt{ms}} \\
\cmidrule(lr){2-4} \cmidrule(lr){5-7} \cmidrule(lr){8-10}  \cmidrule(lr){11-13} 
\multicolumn{1}{c}{}                        & BLEU                         & COMET-22                     & XCOMET                      & BLEU                         & COMET-22                     & XCOMET                      & BLEU                           & COMET-22                       & XCOMET                       & BLEU                         & COMET-22                     & XCOMET                      \\ \hline
Llama-3.1                                   & 43.5                         & 84.5                         & 73.7                        & 32.7                         & 78.8                         & 64.6                        & 1.6                            & 47.0                           & 11.2                         & 34.0                         & 86.9                         & 75.9                        \\
NLLB-3.3B                                   & 51.1                         & 88.3                         & 76.9                        & 46.4                         & 91.2                         & 77.9                        & 17.7                           & 81.6                           & 59.9                         & 41.6                         & 89.1                         & 76.8                        \\
\hline
\end{tabular}
}
\end{table}

\begin{table}[h]
\caption{
Full results for Group 5 in the Flores test data.
}
\vskip 0.05in
\label{app:tab:full_group5}
\centering

\resizebox{1\linewidth}{!}{
\begin{tabular}{lcccccccccccc}
\hline
\multicolumn{1}{c}{\multirow{2}{*}{Models}} & \multicolumn{3}{c}{\texttt{en}$\rightarrow$\texttt{cs}} & \multicolumn{3}{c}{\texttt{en}$\rightarrow$\texttt{el}} & \multicolumn{3}{c}{\texttt{en}$\rightarrow$\texttt{hu}}      & \multicolumn{3}{c}{\texttt{en}$\rightarrow$\texttt{lt}} \\
\cmidrule(lr){2-4} \cmidrule(lr){5-7} \cmidrule(lr){8-10}  \cmidrule(lr){11-13} 
\multicolumn{1}{c}{}                        & BLEU                         & COMET-22                     & XCOMET                      & BLEU                         & COMET-22                     & XCOMET                      & BLEU                           & COMET-22                       & XCOMET                       & BLEU                         & COMET-22                     & XCOMET                      \\ \hline
Llama-3.1                                   & 27.1                         & 88.0                         & 73.5                        & 19.7                         & 82.3                         & 70.4                        & 19.8                           & 82.2                           & 70.4                         & 13.0                         & 77.2                         & 60.6                        \\
NLLB-3.3B                                   & 32.2                         & 91.0                         & 77.7                        & 27.4                         & 89.0                         & 76.6                        & 26.4                           & 89.3                           & 78.7                         & 25.2                         & 89.3                         & 77.3                        \\
\hline
\end{tabular}
}
\end{table}

\begin{table}[h]
\caption{
Full results for Group 6 in the Flores test data.
}
\vskip 0.05in
\label{app:tab:full_group6}
\centering

\resizebox{1\linewidth}{!}{
\begin{tabular}{lcccccccccccc}
\hline
\multicolumn{1}{c}{\multirow{2}{*}{Models}} & \multicolumn{3}{c}{\texttt{en}$\rightarrow$\texttt{et}} & \multicolumn{3}{c}{\texttt{en}$\rightarrow$\texttt{fi}} & \multicolumn{3}{c}{\texttt{en}$\rightarrow$\texttt{ja}}      & \multicolumn{3}{c}{\texttt{en}$\rightarrow$\texttt{ka}} \\
\multicolumn{1}{c}{}                        & BLEU                         & COMET-22                     & XCOMET                      & BLEU                         & COMET-22                     & XCOMET                      & BLEU                           & COMET-22                       & XCOMET                       & BLEU                         & COMET-22                     & XCOMET                      \\ \hline
Llama-3.1                                   & 9.3                          & 66.3                         & 48.1                        & 16.0                         & 82.0                         & 70.4                        & 10.8                           & 66.0                           & 25.0                         & 7.0                          & 73.1                         & 48.1                        \\
NLLB-3.3B                                   & 25.0                         & 90.5                         & 79.7                        & 24.1                         & 91.7                         & 81.1                        & 22.6                           & 87.9                           & 75.1                         & 14.8                         & 84.6                         & 70.2                        \\
LLaMAX2-Alpaca-7B                           & 19.0                         & 88.4                         & 76.1                        & 18.4                         & 90.2                         & 79.6                        & 28.1                           & 88.9                           & 78.1                         & 10.7                         & 83.0                         & 68.2                        \\
\hline
\end{tabular}
}
\end{table}

\begin{table}[h]
\caption{
Full results for Group 7 in the Flores test data.
}
\vskip 0.05in
\label{app:tab:full_group7}
\centering

\resizebox{1\linewidth}{!}{
\begin{tabular}{lcccccccccccc}
\hline
\multicolumn{1}{c}{\multirow{2}{*}{Models}} & \multicolumn{3}{c}{\texttt{en}$\rightarrow$\texttt{gu}} & \multicolumn{3}{c}{\texttt{en}$\rightarrow$\texttt{hi}}      & \multicolumn{3}{c}{\texttt{en}$\rightarrow$\texttt{mr}} & \multicolumn{3}{c}{\texttt{en}$\rightarrow$\texttt{ne}} \\
\cmidrule(lr){2-4} \cmidrule(lr){5-7} \cmidrule(lr){8-10}  \cmidrule(lr){11-13}
\multicolumn{1}{c}{}                        & BLEU                         & COMET-22                     & XCOMET                      & BLEU                           & COMET-22                       & XCOMET                       & BLEU                         & COMET-22                     & XCOMET                      & BLEU                         & COMET-22                     & XCOMET                      \\ \hline
Llama-3.1                                   & 11.9                         & 81.7                         & 55.7                        & 21.7                           & 69.6                           & 53.5                         & 7.6                          & 62.4                         & 49.8                        & 4.3                          & 59.6                         & 47.5                        \\
NLLB-3.3B                                   & 24.3                         & 87.2                         & 66.2                        & 34.4                           & 80.9                           & 67.7                         & 17.1                         & 74.3                         & 65.6                        & 16.4                         & 76.5                         & 76.8                        \\
\hline
\end{tabular}
}
\end{table}

\begin{table}[h]
\caption{
Full results for Group 8 in the Flores test data.
}
\vskip 0.05in
\label{app:tab:full_group8}
\centering

\resizebox{1\linewidth}{!}{
\begin{tabular}{lcccccccccccc}
\hline
\multicolumn{1}{c}{\multirow{2}{*}{Models}} & \multicolumn{3}{c}{\texttt{en}$\rightarrow$\texttt{ar}}      & \multicolumn{3}{c}{\texttt{en}$\rightarrow$\texttt{az}} & \multicolumn{3}{c}{\texttt{en}$\rightarrow$\texttt{fa}} & \multicolumn{3}{c}{\texttt{en}$\rightarrow$\texttt{he}} \\
\cmidrule(lr){2-4} \cmidrule(lr){5-7} \cmidrule(lr){8-10}  \cmidrule(lr){11-13}
\multicolumn{1}{c}{}                        & BLEU                           & COMET-22                       & XCOMET                       & BLEU                         & COMET-22                     & XCOMET                      & BLEU                         & COMET-22                     & XCOMET                      & BLEU                         & COMET-22                     & XCOMET                      \\ \hline
Llama-3.1                                   & 18.9                           & 82.9                           & 70.0                         & 6.6                          & 69.6                         & 50.7                        & 21.4                         & 86.1                         & 77.3                        & 20.8                         & 85.3                         & 73.5                        \\
NLLB-3.3B                                   & 27.5                           & 86.3                           & 75.2                         & 14.0                         & 86.9                         & 76.6                        & 22.6                         & 86.5                         & 77.5                        & 30.4                         & 87.8                         & 76.1                        \\
\hline
\end{tabular}
}
\end{table}

\begin{table}[h]
\caption{
Full results for all languages in the WMT'23 test data.
}
\vskip 0.05in
\label{app:tab:full_wmt23}
\centering

\resizebox{1\linewidth}{!}{
\begin{tabular}{lcccccccccccc}
\hline
\multicolumn{1}{c}{\multirow{2}{*}{Models}} & \multicolumn{3}{c}{\texttt{en}$\rightarrow$\texttt{de}} & \multicolumn{3}{c}{\texttt{en}$\rightarrow$\texttt{zh}} & \multicolumn{3}{c}{\texttt{en}$\rightarrow$\texttt{ja}}      & \multicolumn{3}{c}{\texttt{en}$\rightarrow$\texttt{ru}} \\
\cmidrule(lr){2-4} \cmidrule(lr){5-7} \cmidrule(lr){8-10}  \cmidrule(lr){11-13}
\multicolumn{1}{c}{}                        & BLEU                         & COMET-22                     & XCOMET                      & BLEU                         & COMET-22                     & XCOMET                      & BLEU                           & COMET-22                       & XCOMET                       & BLEU                         & COMET-22                     & XCOMET                      \\
ALMA-R-13B                                  & 30.4                         & 84.0                         & 68.8                        & 32.3                         & 85.0                         & 71.3                        & -                              & -                              & -                            & 22.8                         & 85.5                         & \textbf{74.4}               \\
TowerInstruct-7B-v0.2                       & 37.9                         & 83.1                         & 68.1                        & 41.9                         & 85.6                         & 70.2                        & -                              & -                              & -                            & 29.2                         & 85.3                         & 71.1                        \\
\hline
\end{tabular}
}
\end{table}

\clearpage
\section{Examples of Over-Rejection}
\label{app:sec:over_reject}
\begin{figure}[h]
    \centering
    \resizebox{1\linewidth}{!}{
    \includegraphics[width=7.5cm]{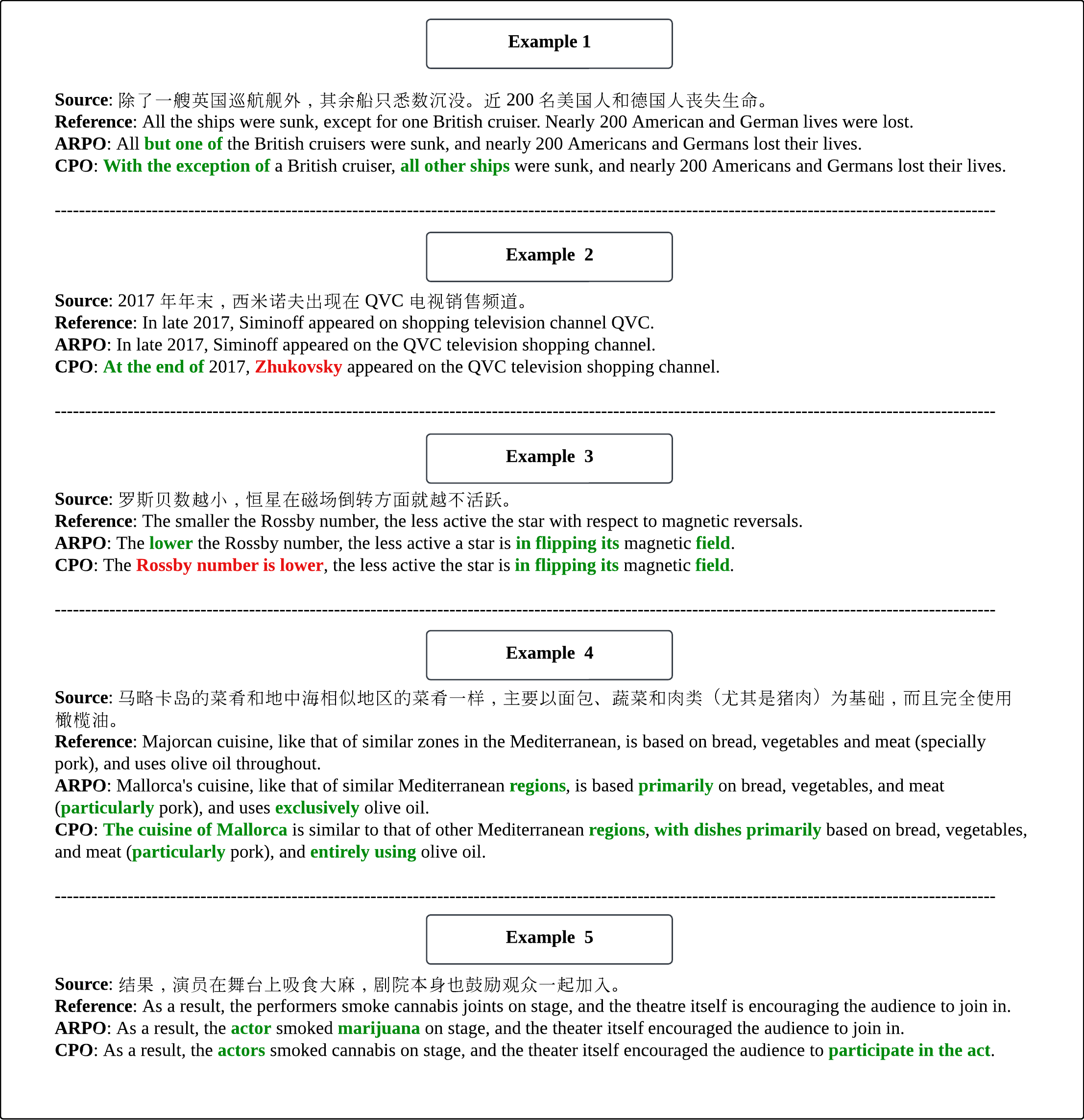}}
    \caption{
     Examples of over-rejection in Chinese-to-English translation, comparing translations from the reference, ARPO, and CPO. \textbf{\green{Green}} highlights indicate acceptable variations, while \textbf{\red{red}} highlights show the potentially harmful changes. CPO introduces more stylistic differences than ARPO, with most being correct but some leading to over-rejection. Although most of variations are correct, the phenomenon of excessive stylistic changes leading to non-optimal translations is referred to as `over-rejection'.
    }
    \label{app:fig:over_reject}
\end{figure}

Figure \ref{app:fig:over_reject} presents examples of over-rejection in translations from Chinese to English. For each source sentence, we provide translations from the reference, ARPO (implemented on CPO), and CPO. The words where ARPO and CPO differ from the reference are color-highlighted: green indicates that the variation does not affect the meaning, while red indicates a potentially negative impact on translation quality. As shown in Figure \ref{app:fig:over_reject}, CPO exhibits more stylistic variations than ARPO across all translation examples. Although most of the stylistic changes introduced by CPO are accurate and do not impair meaning, a small number are detrimental. Excessive changes in style can result in sub-optimal translations, a phenomenon we refer to as `over-rejection'.

\end{document}